\documentclass[journal]{IEEEtran}

\usepackage{amsmath}
\usepackage{amssymb}
\usepackage{epsfig}
\usepackage{graphicx}
\usepackage{graphics}
\usepackage{multirow}
\usepackage{tabu}

\ifCLASSINFOpdf

\else

\fi

\begin{document}

\title{Movable-Object-Aware Visual SLAM via Weakly Supervised Semantic Segmentation}

\author{Ting~Sun$^{1}$, Yuxiang~Sun$^{1}$, Ming~Liu$^{1}$, Dit-Yan~Yeung$^{2}$
	\thanks{$^{1}$ Department of Electronic \& Computer Engineering, Hong Kong University of Science and Technology, Hong Kong, China. {\tt\small (tsun, eeyxsun, eelium)@ust.hk} }
	\thanks{$^{2}$ Department of Computer Science, Hong Kong University of Science and Technology, Hong Kong, China. {\tt\small dyyeung@cse.ust.hk}}%
}


\maketitle

\begin{abstract}
Moving objects can greatly jeopardize the performance of a visual simultaneous localization and mapping (vSLAM) system which relies on the static-world assumption.  Motion removal have seen successful on solving this problem.  Two main streams of solutions are based on either geometry constraints or deep semantic segmentation neural network.  The former rely on static majority assumption, and the latter require labor-intensive pixel-wise annotations.  In this paper we propose to adopt a novel weakly-supervised semantic segmentation method.  The segmentation mask is obtained from a CNN pre-trained with image-level class labels only.   Thus, we leverage the power of deep semantic segmentation CNNs, while avoid requiring expensive annotations for training.  We integrate our motion removal approach with the ORB-SLAM2 system. Experimental results on the TUM RGB-D and the KITTI stereo datasets demonstrate our superiority over the state-of-the-art.
\end{abstract}


\IEEEpeerreviewmaketitle

\section{Introduction}
\label{sec:introduction}
Visual simultaneous localization and mapping (vSLAM) is widely adopted by robots to concurrently estimate its poses and reconstruct the traversed environments using visual sensors, such as monocular
cameras~\cite{younes2017keyframe}, stereo cameras~\cite{gomez2019pl} and RGB-D cameras~\cite{whelan2015real}.  Over the past decades, many well-performing SLAM systems have been developed such as SVO~\cite{Forster2014ICRA}, LSD-SLAM~\cite{engel2014lsd} and ORB-SLAM2~\cite{murORB2}, and most of them adopt the graph optimization framework~\cite{cadena2016past}.  These approaches build a graph whose nodes correspond to the poses of the robot at different points and whose edges represent the constraints between the poses.  The edges are obtained from the observations of the environment or from movement actions carried out by the robot~\cite{grisetti2010tutorial}.  The graph is built in the front end and is optimized in the back end to find the configuration that is most consistent with the measurements.  

The vast majority of existing vSLAM systems heavily rely on static-world assumption, but this is hardly true for many application scenarios such as autonomous driving.  The moving objects hinder the data associations in both short-term and long-term \cite{cadena2016past}.  In the sort-term, the adjacent pose estimated w.r.t. moving landmarks is inaccurate; in the long-term, the loop detection would be confused by matching the same scene with different objects layout.

In recent years, there are emerging technologies to make the existing vSLAM systems moving-object-aware~\cite{sun2017improving,li2017rgb,f2018detect,yao2018robust,bescos2018dynaslam,sun2018motion,yu2018ds,brasch2018semantic,fan2018dynamic,xiao2019dynamic}.  Most of these methods modify the front end to prevent the moving objects being treated as landmarks, saving the trouble to alter the back end.  Besides the objects that are currently moving, we think it is reasonable to also detect the potentially movable objects, as they may harm the loop detection in the long-term.

There are two main assumptions used in existing motion removal methods: 1) the majority regions are static, and 2) the movable objects belong to certain known categories.  The first assumption justifies the usage of some standard robust estimators, such as the RANdom SAmple Consensus (RANSAC) algorithm~\cite{fischler1981random}, but fails when moving objects cover a major part of the camera field of view.  The second assumption incorporates semantic prior knowledge, which is usually true for the specific environment in which the robot operates, and it also takes care of the temporal static movable objects, preventing its harm to long-term data-association.

Recent decades witness the prosperity of deep learning \cite{deep-learning,deep-learning-overview} technologies. In particular, convolutional neural networks (CNN or ConvNet) \cite{convnet} provide a powerful end-to-end framework which achieve state-of-the-art performance in many challenging computer vision tasks~\cite{my_fine_grained}.  A semantic segmentation CNN takes a color image as input, and output a mask which labels every pixel in the image with one of the several predetermined categories.  This segmentation mask can be easily used in the vSLAM systems to precisely separate the movable object region and the static background region.  However, to train a deep CNN requires a large amount of data.  Though there are a lot of images available, but the pixel-level annotations are laborious and expensive to collect, and this severely limits the applicability and adaptability of using a segmentation CNN in vSLAM problems. 

In this paper, we propose to use a novel weakly-supervised semantic segmentation method \cite{sun2019fully} to solve this problem.  The network is pre-trained using image-level class labels only, its outputs are refined by conditional random field (CRF) \cite{krahenbuhl2011efficient}.  From the segmentation result, we generate a binary mask which indicates all the pixels belong to movable objects.  We pass this binary mask to the tracking thread of ORB-SLAM2 system, forcing the feature points to avoid movable objects.  This method can be applied to the vSLAM system as long as camera images are used.  In particular, when depth images are available, we incorporate them into the CRF \cite{krahenbuhl2011efficient} refinement of segmentation mask to further improve the performance.  

Our main contribution is that we are the first to propose using a weakly-supervised semantic segmentation CNN \cite{sun2019fully} for vSLAM in dynamic environment.  We modify the ORB-SLAM2 system with proposed method so that it will not use movable objects as landmarks, avoiding erroneous data association in both short-term and long-term.  This method leverages the power of deep CNN, without using expensive annotations for training.  Thus, it is more applicable and adjustable compared with fully-supervised deep learning approaches.  Experimental results on TUM RGB-D \cite{sturm12iros} and stereo KITTI \cite{kitti} dataset demonstrate that our approach significantly improve the ORB-SLAM2 and achieves the state-of-the-art performance in various challenging scenarios.

The remainder of this paper is organized as follows.  Sec.~\ref{sec:related work} reviews previous related works.  The weakly-supervised semantic segmentation method used in our system and how we customize it to leverage depth images for CRF refinement is explained in Sec.~\ref{sec:weak seg}.  Sec.~\ref{sec:modified ORB-SLAM2} gives the details of how we integrate segmentation results into ORB-SLAM2~\cite{murORB2}.  The experimental results are presented in Sec.~\ref{sec:experiments}.  Sec.~\ref{sec:conclusion} concludes this paper.

\section{Related work}
\label{sec:related work}

\subsection{vSLAM for Dynamic Scenes}
As mentioned previously, most of the existing vSLAM systems rely on static-world assumption.  The constructed map contains landmarks at fixed positions, and the robot estimates its own poses w.r.t. them.  Moving objects in the environment will cause both short-term and long-term erroneous data associations.  Recently proposed methods that address this moving-object issue are mainly based on two assumptions: 1) the majority regions are static, and 2) the movable objects belong to certain known categories.  

Y. Sun \textit{et al.} \cite{sun2018motion} propose to first use optical flow to estimate the 2-D homography transformation, then identify moving foreground pixels from the reprojection error and build the foreground model.  In the inference process, they pixel-wisely compare the current RGB-D frame with the model to segment the foreground.  Besides assumption 1), this method also assumes that planes are static objects, and each sequence has to build its won moving foreground model.  S. Li and D. Lee \cite{li2017rgb} use an organized point cloud to detect edge points, and estimate initial camera transformation that align the source cloud to the target cloud.  The initial camera pose is used to weigh each edge point's static property, then Intensity Assisted Iterative Closest Point (IAICP) method \cite{li2016fast} is adopted for refinement with weighted reprojection loss.  This method also relies on assumption 1) and requires depth image, while our method is applicable to all monocular, stereo and RGB-D settings.  Similar approach is proposed in \cite{yao2018robust}, which is based on ORB-SLAM2 \cite{murORB2}, and feature points fall into `dynamic region' are ignored during egomotion estimation, and static weights calculated by distance transform errors are added to pose estimation. 

Some recent approaches propose to leverage deep learning models based on assumption 2), in particular, deep segmentation CNNs take raw image as input, and output pixel-wise class labels which can be easily used in vSLAM.  In \cite{bescos2018dynaslam}, B. Bescos \textit{et al.} use Mask R-CNN \cite{he2017mask} for movable object detection, in \cite{yu2018ds} C. Yu \textit{et al.} use SegNet \cite{badrinarayanan2017segnet} and in \cite{brasch2018semantic} N. Brasch \textit{et al.} use ICNet \cite{zhao2018icnet}.  However, all Mask R-CNN \cite{he2017mask}, SegNet \cite{badrinarayanan2017segnet} and ICNet \cite{zhao2018icnet} are trained in a fully-supervised manner, if the robots are to operate in an environment with new types of movable objects, it will be very expensive to collect a dataset with pixel-wise annotation.  Thus, their adaptability is limited.   L. Xiao \textit{et al.} \cite{xiao2019dynamic} use SSD \cite{liu2016ssd} with prior knowledge for movable objects detection.  However, the object detector outputs a bounding box rather than pixel-wise labels, and this is too coarse that a large portion of feature points in the static background close to the movable objects are mistakenly ignored.

There are two common pipelines adopted by the above mentioned methods to modify ORB-SLAM2 \cite{murORB2}: 1) pre-processing the input image e.g. masking or inpainting the moving object region in the input image \cite{bescos2018dynaslam}; 2) treating the feature points fall into the detected movable object region as outliers \cite{bescos2018dynaslam,yu2018ds,xiao2019dynamic} and removing them before data association.  The first approach may introduce some artifact and fake feature points, the second approach will cause big variation in the number of feature points used in different frames.  Our implementation passes the movable object mask into the ORB feature points extraction module, and keeps a stable number of feature points used for each frame.

\subsection{Weakly Supervised Semantic Segmentation} 
Weakly supervised semantic segmentation has been extensively studied to relieve the data deficiency problem.  According to the types of annotation required by the overall system, existing weakly-supervised methods are based on various annotations such as user scribble \cite{tang2018normalized}, web images \cite{shen2018bootstrapping}, bounding box  \cite{dai2015boxsup} etc.  Since image level annotation, i.e., class labels, are abundant and relatively cheap to collect, we adopt a segmentation system \cite{sun2019fully} that only requires image-level class labels during training.  This method first generates segmentation cues from two classifiers, then uses them to train a segmentation network.  The segmentation result are refined by CRF \cite{krahenbuhl2011efficient} at last.

\section{Adopted weakly supervised semantic segmentation}
\label{sec:weak seg}
Training a segmentation network in a fully supervised manner requires a large amount of pixel-wise annotations which are laborious and expensive to collect.  So weakly-supervised semantic segmentation is receiving growing attention and will have a significant impact on this area.  Among various types of annotations, image level class labels are very cheap to collect, so we adopt a lately proposed weakly-supervised semantic segmentation system \cite{sun2019fully}, which is trained by image-level class labels only. 

\begin{figure*}[h]
	\centering
	\includegraphics[width=0.9\textwidth]{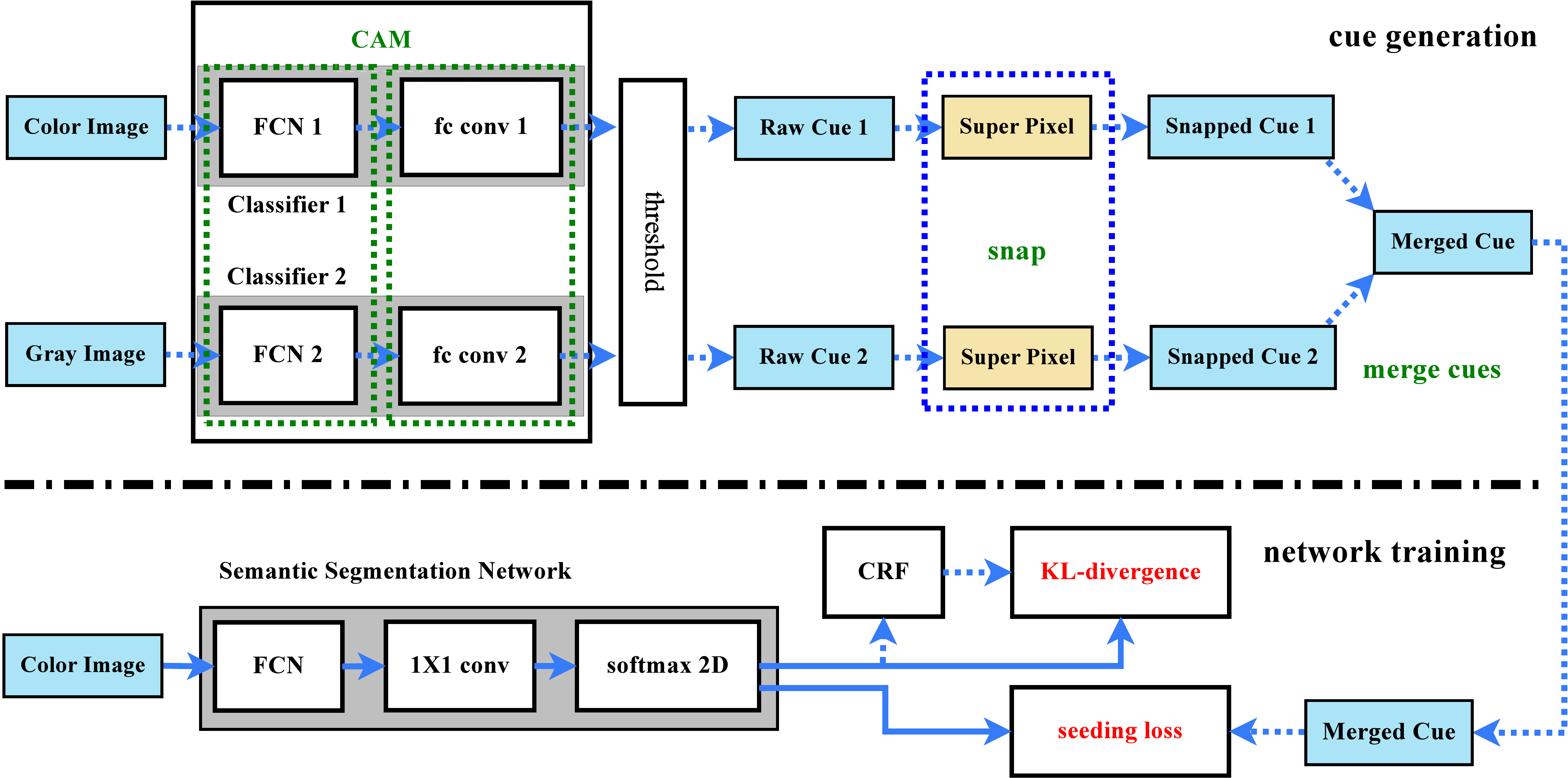}
	\caption{The overview of the adopted weakly supervised semantic segmentation system proposed in \cite{sun2019fully}.  The top half shows how to generate cues from two classifiers.  The bottom half shows how to use the cues to train a segmentation network.  During the testing phase, the output of CRF are used as prediction, and the two loss term, i.e. KL-divergence and seeding loss are ignored.  The overall system only require image-level class labels to train the classifiers.  More details can be found in \cite{sun2019fully}.}
	\label{fig:adopted seg sys}
	\vspace{-5mm}
\end{figure*}

T. Sun \textit{et al.} \cite{sun2019fully} follows a common pipeline of weakly supervised semantic segmentation, i.e. they first obtain `pseudo annotations', then use these `pseudo annotations' to train a segmentation CNN.  The overview of our adopted segmentation system \cite{sun2019fully} is shown in Figure~\ref{fig:adopted seg sys}.  The class-specific activation maps from the classifiers are used as cues to train a segmentation network. The well-known defects of these cues are coarseness and incompleteness.  T. Sun \textit{et al.} \cite{sun2019fully} use super-pixel to refine them, and fuse the cues extracted from both a color image trained classifier and a gray image trained classifier to compensate for their incompleteness.  The CRF is adapted to regulate the training process and to refine the outputs. More details can be found in \cite{sun2019fully}.

In RGB-D SALM setting, the depth images help to separate the objects at different distances from the camera.  By using these depth images, we add another customized bilateral term in the pairwise potential.  The formulation of this term is the same as that of the color image, but the color value is replaced by the depth value at pixel location.

In \cite{sun2019fully}, the segmentation CNN is trained on PASCAL VOC 2012 augmented dataset \cite{everingham2010pascal,mark2015ijcv,hariharan2011semantic}.  We pick the following classes as movable object types: aeroplane, bike, bus, car, motorbike and person.

\section{Modified ORB-SLAM2}
\label{sec:modified ORB-SLAM2}
\begin{figure*}[h]
	\centering
	\includegraphics[width=\textwidth]{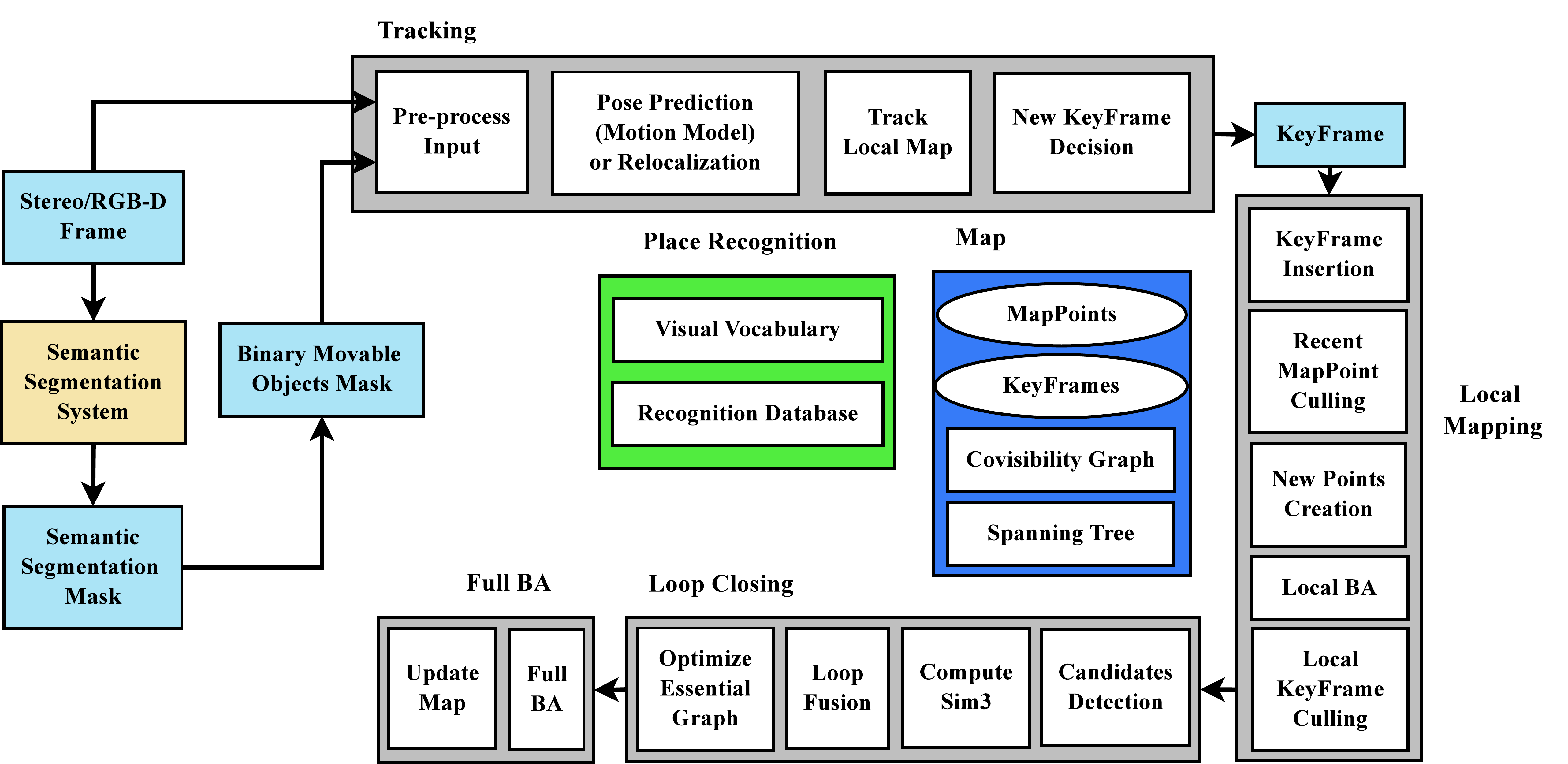}
	\caption{The ORB-SLAM2 system \cite{murORB2} modified with a weakly supervised semantic segmentation system.}
	\label{fig:modified ORB-SLAM2}
	\vspace{-3mm}
\end{figure*}

The modified ORB-SLAM2 system \cite{murORB2} is shown in Figure~\ref{fig:modified ORB-SLAM2}, where the \textit{Semantic Segmentation System} is composed of a CNN followed by CRF.  Original ORB-SLAM2 \cite{murORB2} directly takes \textit{Stereo/RGB-D Frame} as input.  In the proposed system, the color images first go through the semantic segmentation system to produce corresponding semantic segmentation masks, from which the \textit{Binary Movable Object Masks} are generated.  In these \textit{Binary Movable Object Masks}, all the pixels whose labels belong to one of the predefined movable object classes are indicated by 1.  Both the original images and the binary masks are input to the tracking thread of ORB-SLAM2 \cite{murORB2}, and the detailed implementation is shown in Figure~\ref{fig:modified ORB-extractor}, where some variable names are taken from the code of ORB-SLAM2 \cite{murORB2}.  The modules in the ORB feature extraction process are shown in white rectangles and the data are blue.  The \textit{Binary Movable Object Mask} is colored green since it is the new input we pass to \textit{ORBExtractor}.  The points labeled 1 in \textit{Binary Movable Object Mask} will not be buffered in \textit{ToDistributeKeys}.  The number of key points in a frame is controlled in \textit{DistributeOctTree}.

There are several merits of our implementation:

\begin{figure*}[h]
	\centering
	\includegraphics[width=\textwidth]{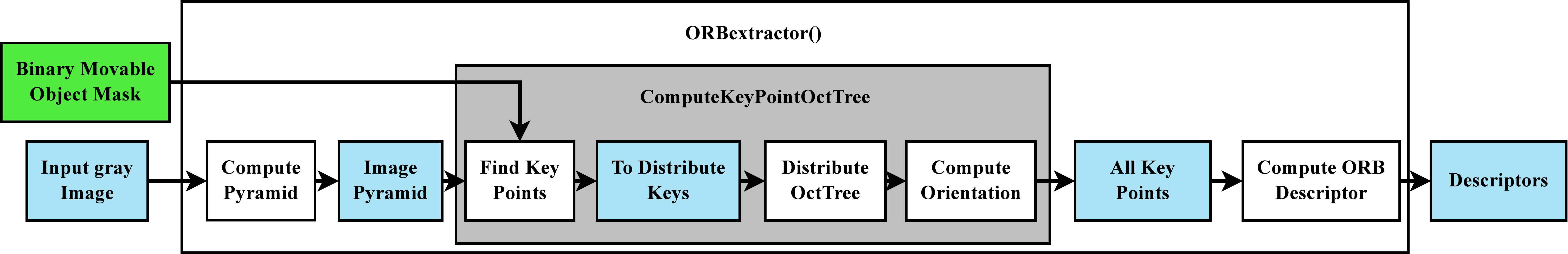}
	\caption{The modified ORB feature extraction process. Detailed description can be found in the text.}
	\label{fig:modified ORB-extractor}
	\vspace{-5mm}
\end{figure*}

\begin{itemize}
	\item Compared with masking or inpainting the input image as a preprocess before \textit{`tracking'} thread, our method does not introduce artifacts that may create erroneous feature points, and we save the computation for preprocessing the input images.
	\item Compared with treating the feature points fall into the movable object region as outliers and ignore them, our method maintains the fixed number of feature points to be detected in each frame.
\end{itemize}

\begin{figure*}[h]
	\centering
	\includegraphics[width=0.19\textwidth]{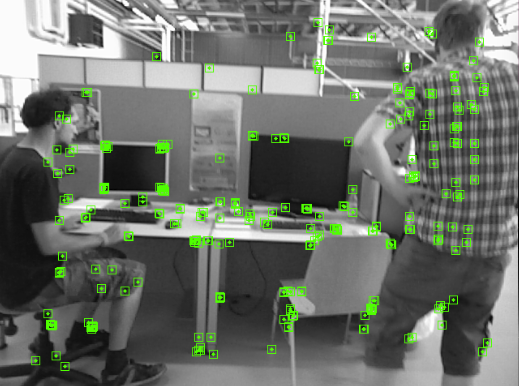}
	\includegraphics[width=0.19\textwidth]{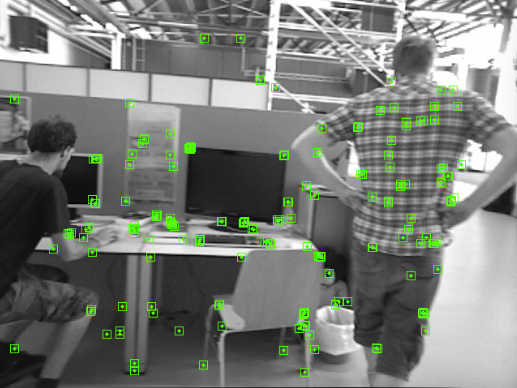}
	\includegraphics[width=0.19\textwidth]{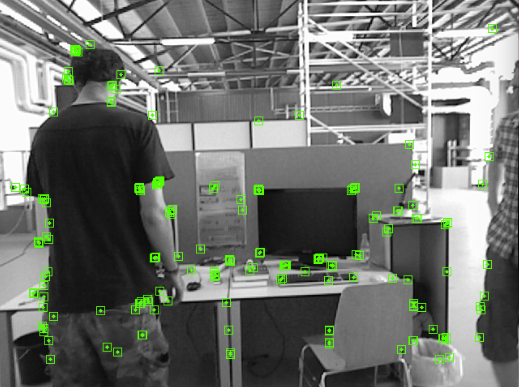}
	\includegraphics[width=0.19\textwidth]{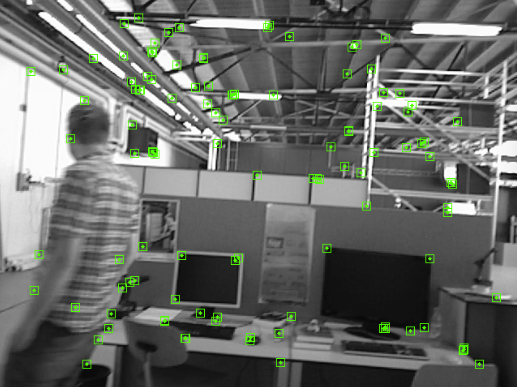}
	\includegraphics[width=0.19\textwidth]{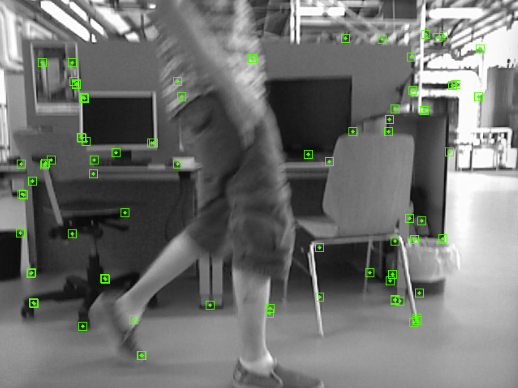} \\
	\includegraphics[width=0.19\textwidth]{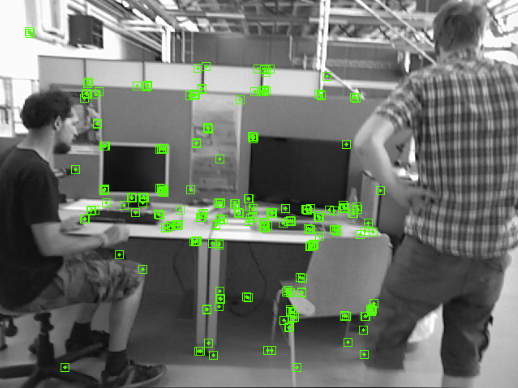}
	\includegraphics[width=0.19\textwidth]{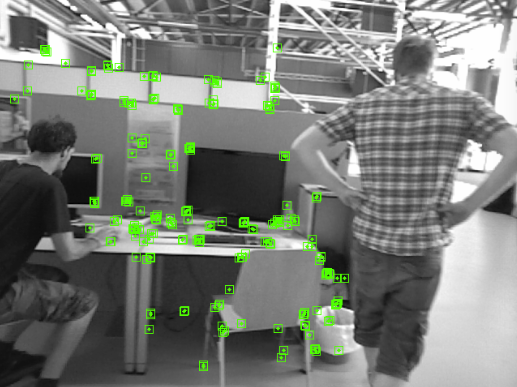}
	\includegraphics[width=0.19\textwidth]{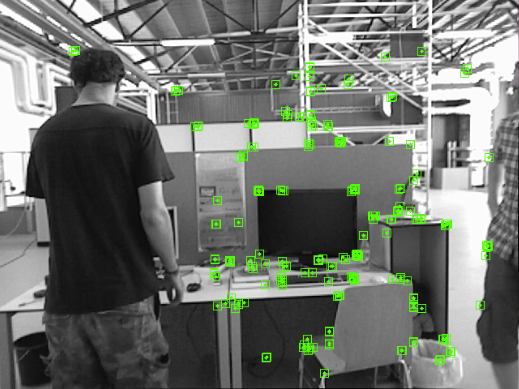}
	\includegraphics[width=0.19\textwidth]{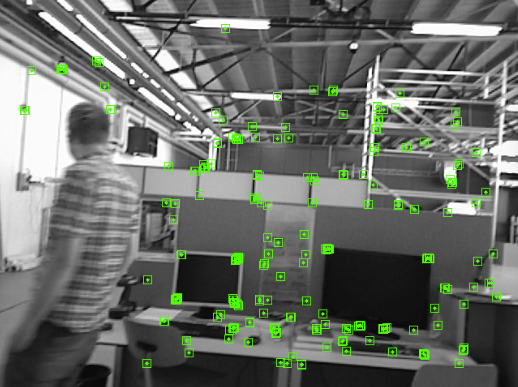}
	\includegraphics[width=0.19\textwidth]{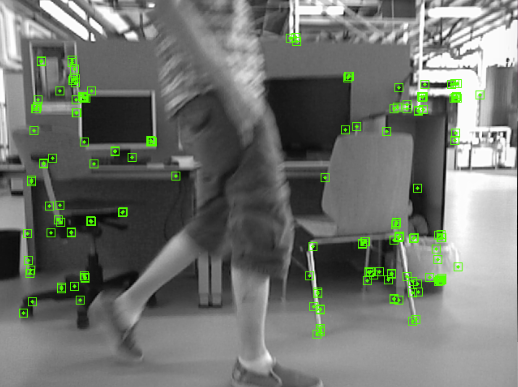} 
	\caption{Sample frames marked with detected feature points.  The first row shows the feature points detected by ORB-SLAM2 \cite{murORB2}, and the second row shows the feature points detected by our modified ORB-SLAM2 \cite{murORB2}.  It can be seen that in our proposed system, the feature points do not fall into the movable object regions.}
	\label{fig:sample frames}
	\vspace{-3mm}
\end{figure*}

Some sample frames marked with detected feature points are shown in Figure~\ref{fig:sample frames}.  The first row shows the feature points detected by ORB-SLAM2 \cite{murORB2}, and the second row shows the feature points detected by our modified ORB-SLAM2 \cite{murORB2}.  It can be seen that in our proposed system, the feature points do not fall into the movable object regions.

\section{Experiments}
\label{sec:experiments}
\subsection{Experiment setup}
Our method is implemented based on ORB-SLAM2 \cite{murORB2} with its default parameters, and is tested on two datasets: the Dynamic Objects sequences of TUM RGB-D dataset \cite{sturm12iros} and stereo KITTI \cite{kitti} dataset.  The semantic segmentation masks are generated on a desktop PC with 3.30GHz Intel i9-7900X CPU, 46GB RAM, and a GeForce GTX 1080.  Our segmentation CNN is implemented in Pytorch \cite{paszke2017automatic}, and we adopt the public available CRF implementation in \cite{krahenbuhl2011efficient}. 

Here we introduce some shortened names used in this section. For the Dynamic objects sequences of TUM RGB-D dataset \cite{sturm12iros}, we use f,w,s,v for freiburg, walking, sitting, validation respectively, e.g. \textit{f3/w/xyz/v} represents \textit{rgbd\_dataset\_freiburg3\_walking\_xyz\_validation}.  Among these TUM sequences, the sitting sequences depict low-dynamic scenarios, while the walking sequences depict high-dynamic scenarios \cite{sun2017improving}.  As mentioned in Sec.~\ref{sec:weak seg}, CRF is adopted to refine the outputs of the weakly supervised semantic segmentation CNN \cite{sun2019fully}.  For RGB-D case, the pairwise potential of the CRF can be obtained from color image alone, depth image alone or both, and we use `c', `d' and `c,d' to represent these CRF settings respectively.  For example, `+c' represents the configuration: modifing the ORB-SLAM2 with proposed method, whose CRF potentials are defined using color images only, while `+c,d' means the CRF potentials are defined using both color images and depth images.

\subsection{Evaluation metric}
We employed the widely used metrics Absolute Trajectory Error (ATE) and Relative Pose Error (RPE) for the quantitative evaluations \cite{sturm12iros}. The RPE contains both the translational drift error
and the rotational drift error.  We present the values of Root Mean Square Error (RMSE), Mean Error, Median Error and Standard Deviation (S.D.) in this paper. The improvements brought by our approach are calculated using the following formula:

\begin{equation}\label{equ:improvement}
	F = (1 - \frac{\beta}{\alpha}) \times 100 \%
\end{equation}

where \(F\) represents the RMSE improvement value, \(\alpha \) represents the RMSE obtained without our approach, \(\beta \) represents the RMSE obtained with our approach.

\subsection{Results and analysis}
\subsubsection{Results on TUM RGB-D dataset \cite{sturm12iros}}
\begin{figure*}[!h]
	\centering
	\includegraphics[width=0.19\textwidth]{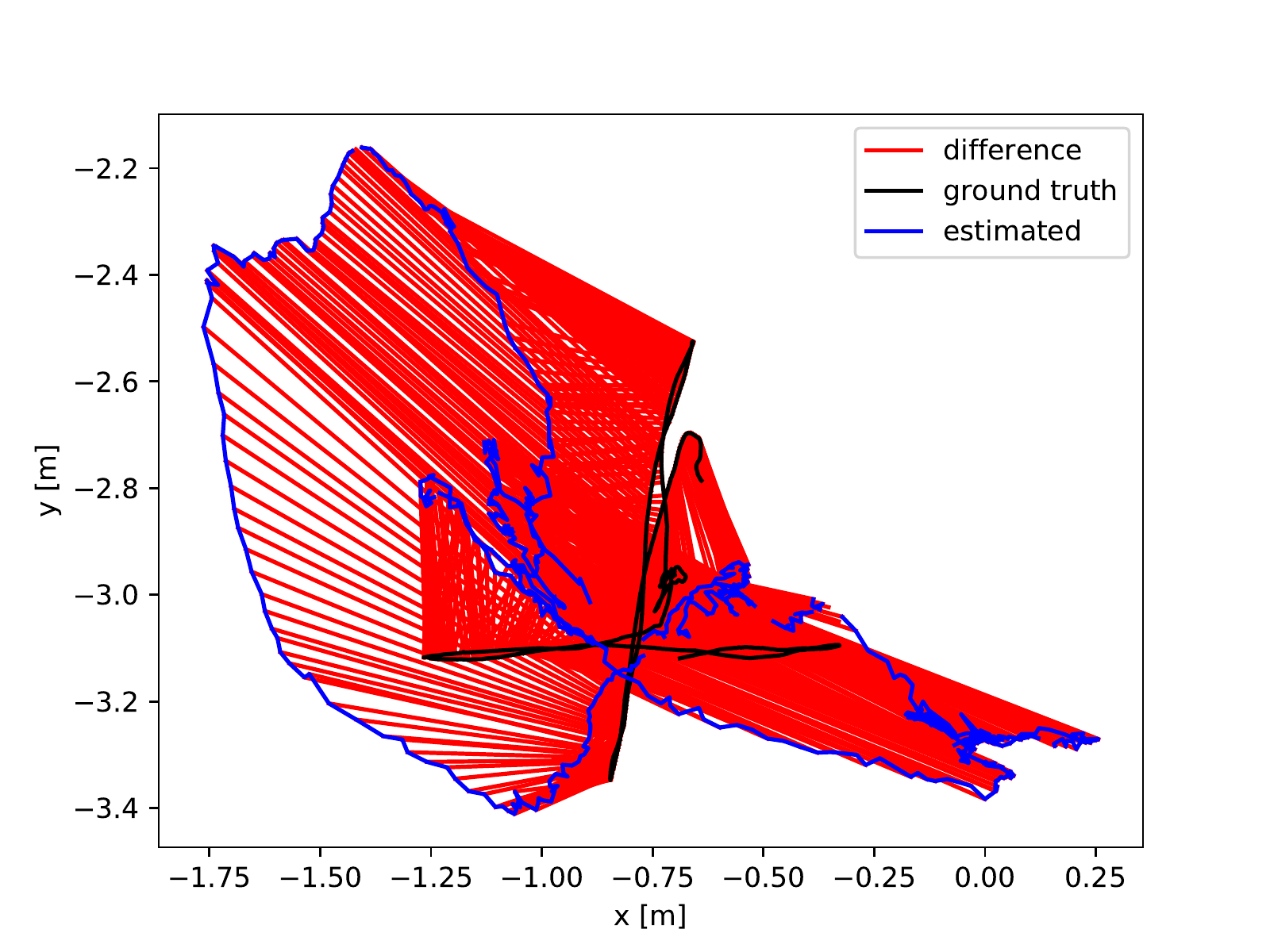}
	\includegraphics[width=0.19\textwidth]{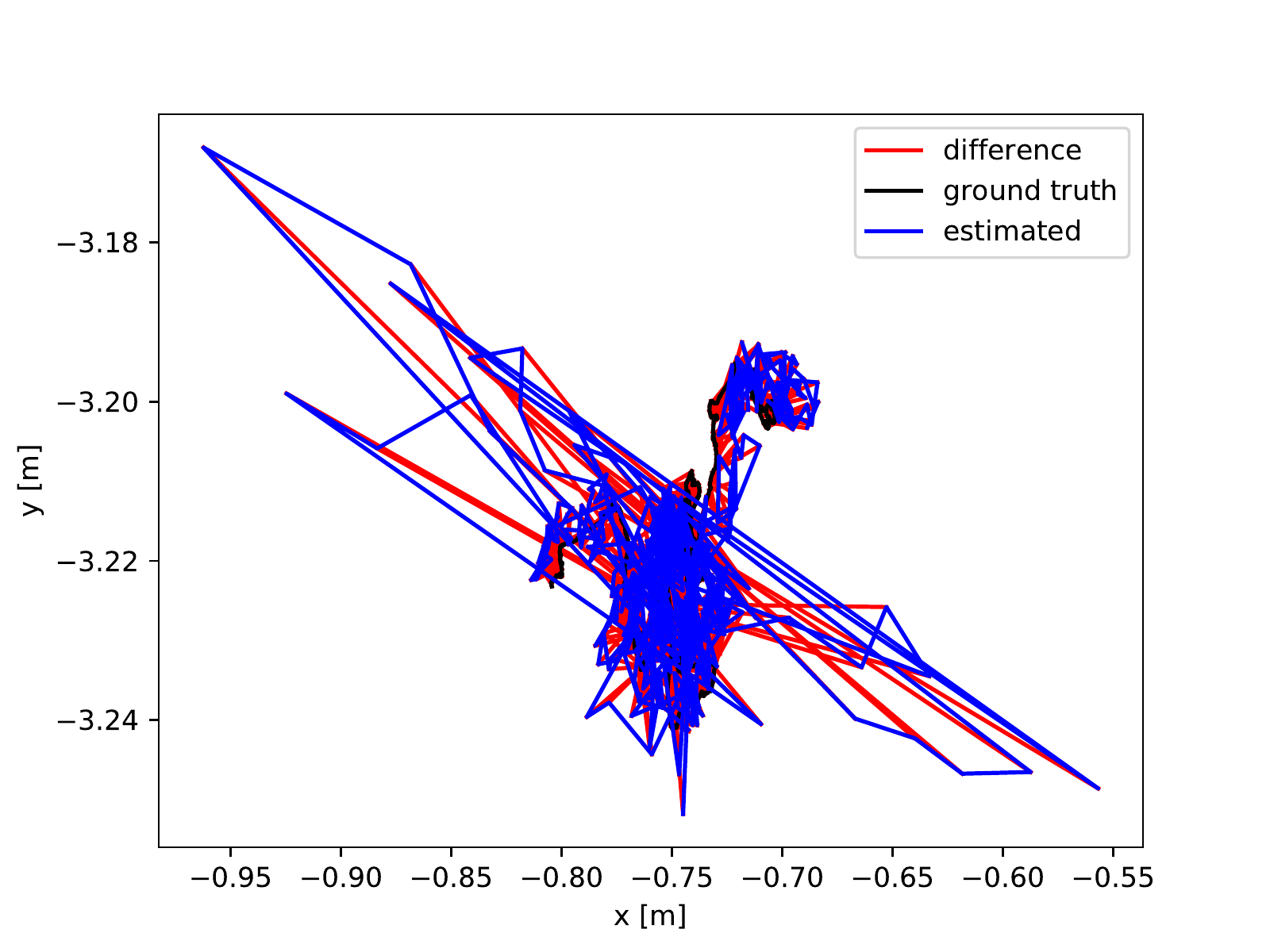}
	\includegraphics[width=0.19\textwidth]{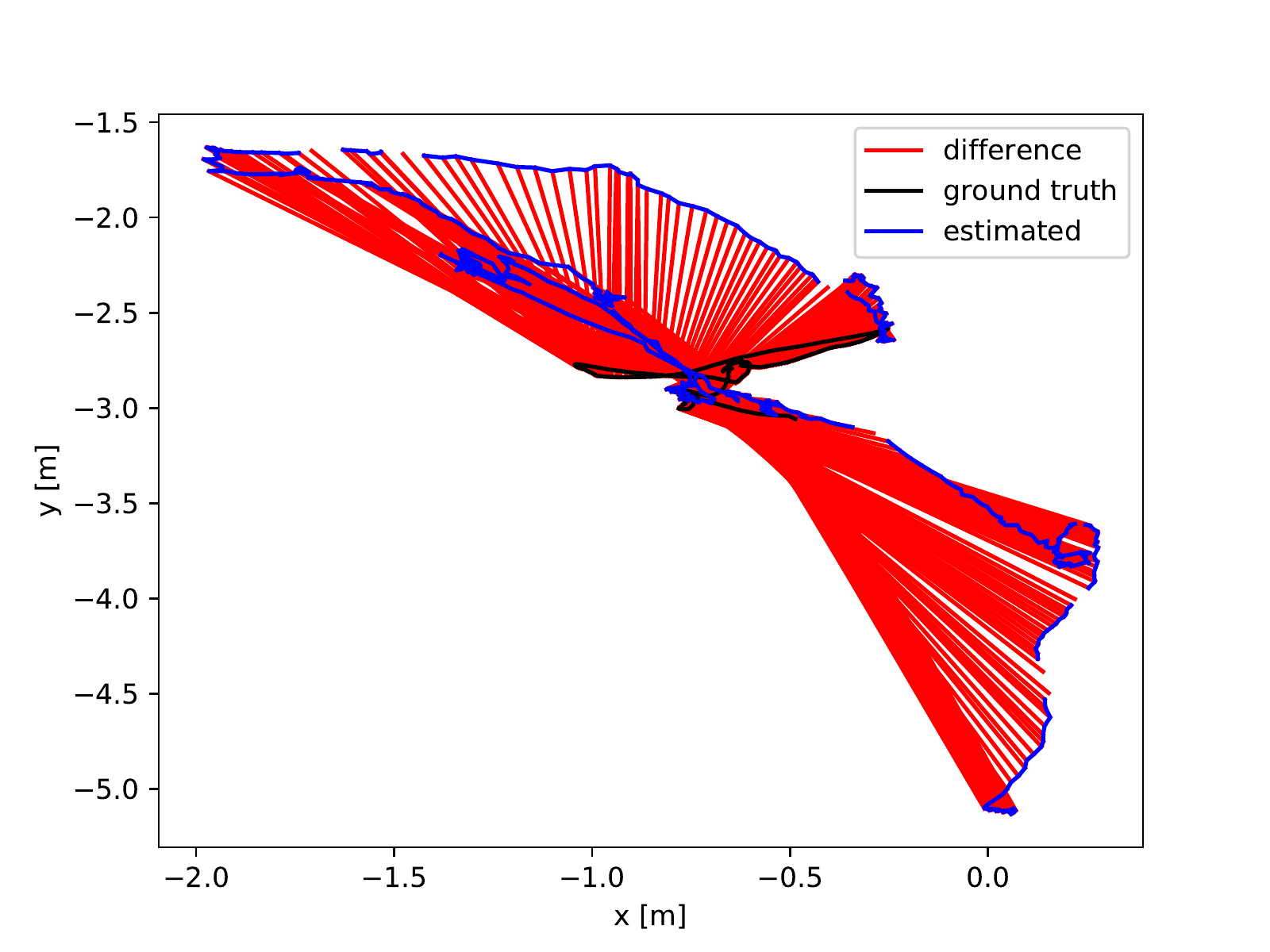}
	\includegraphics[width=0.19\textwidth]{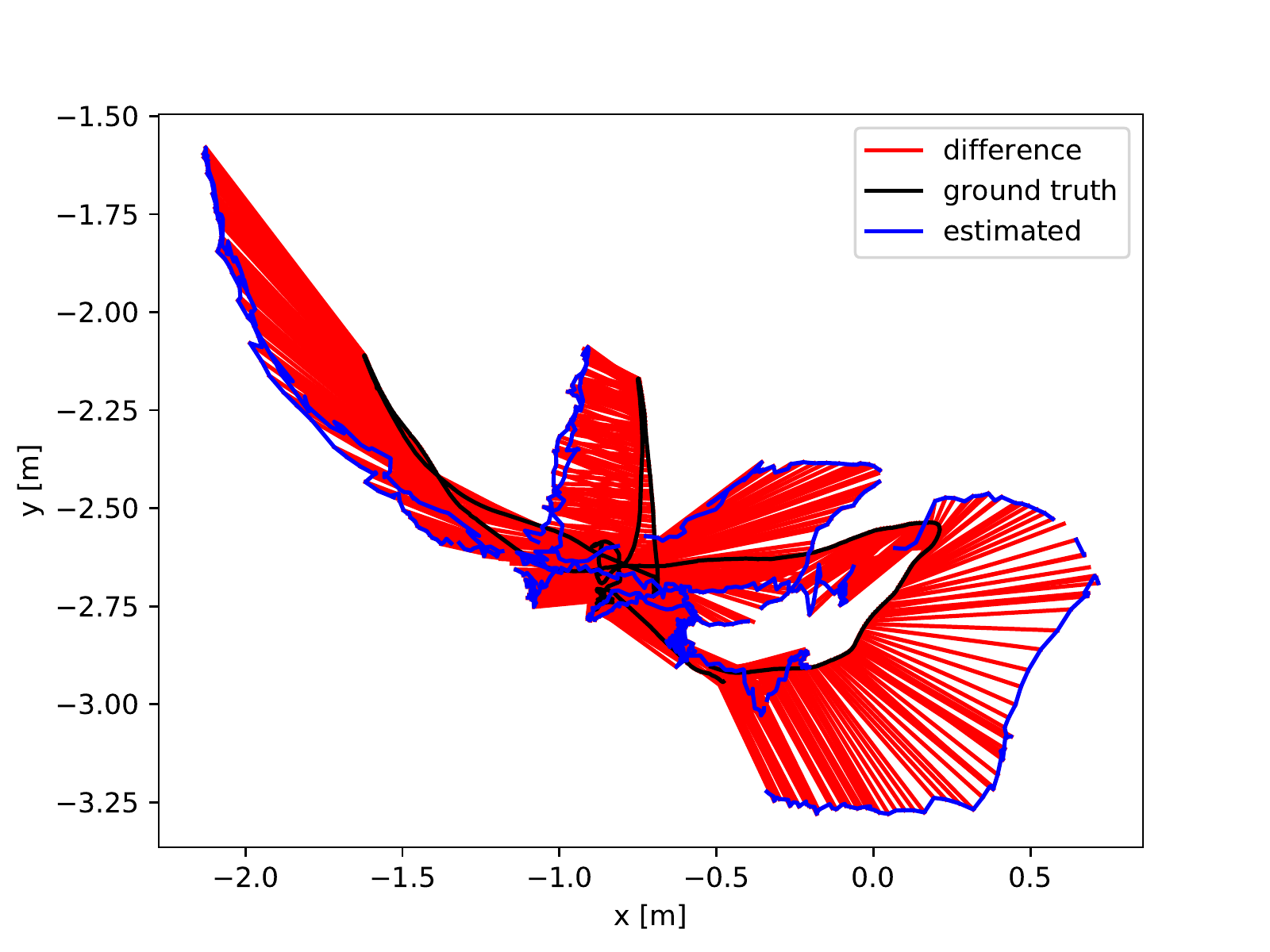}
	\includegraphics[width=0.19\textwidth]{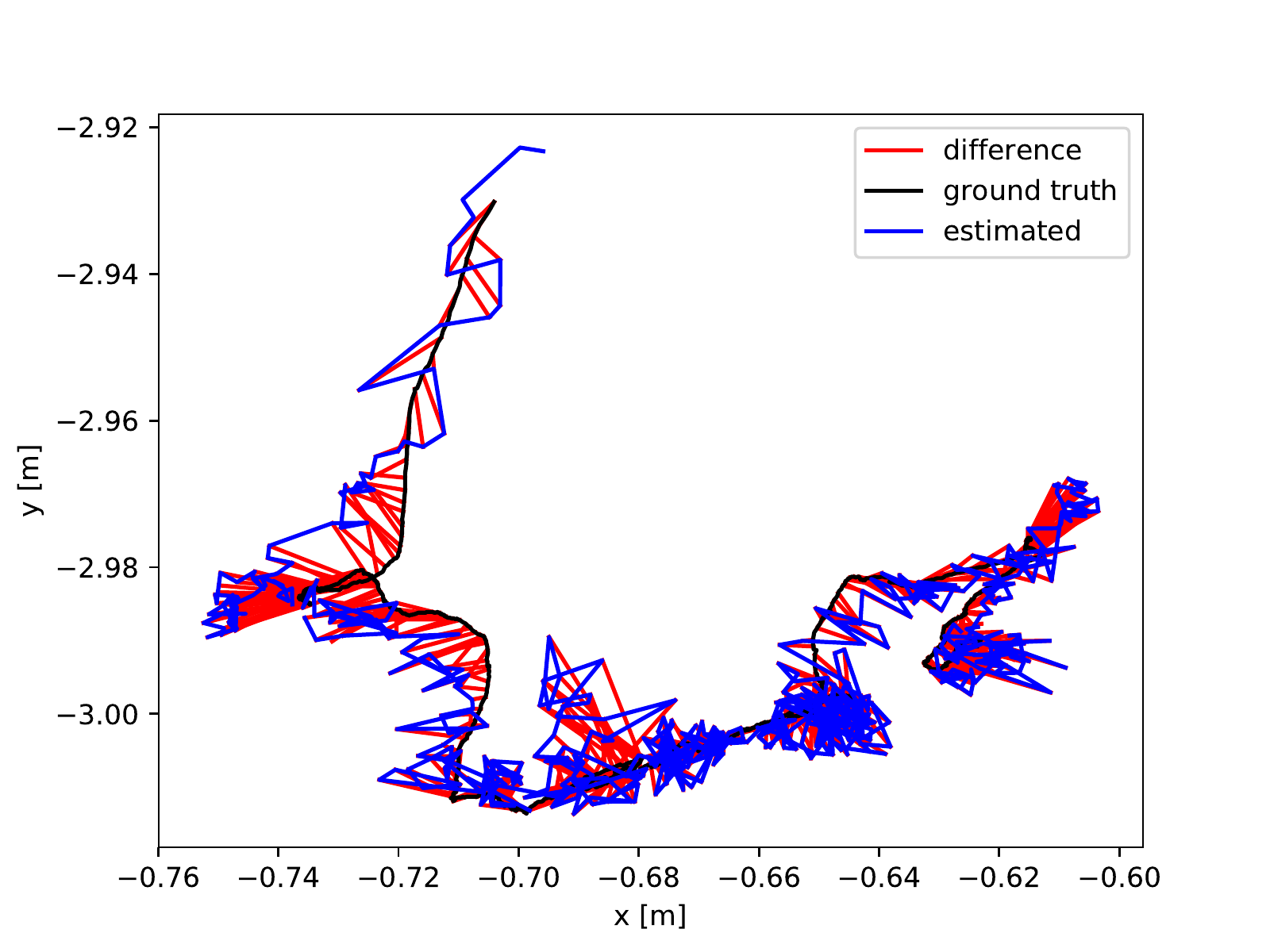} \\
	\includegraphics[width=0.19\textwidth]{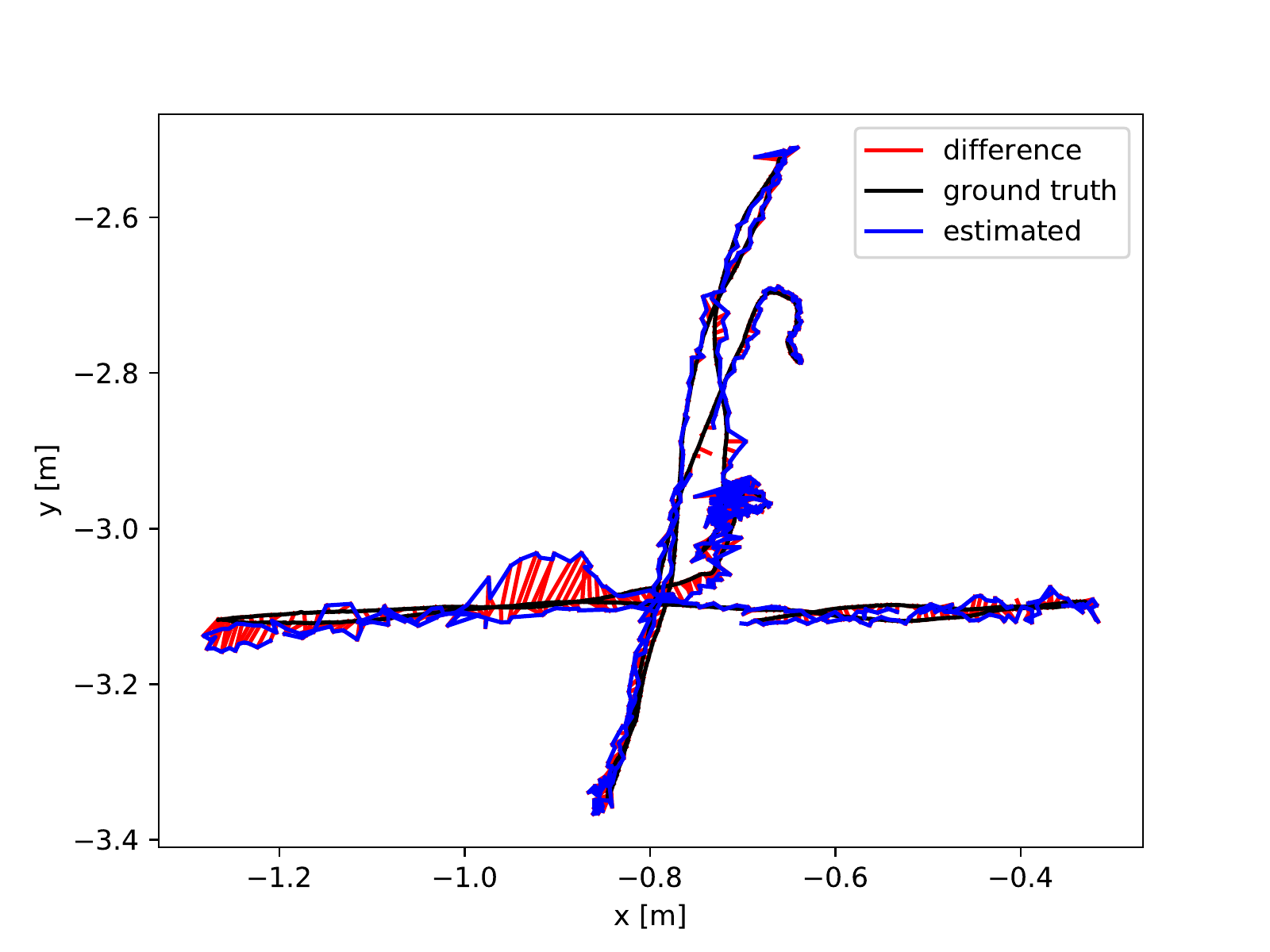}
	\includegraphics[width=0.19\textwidth]{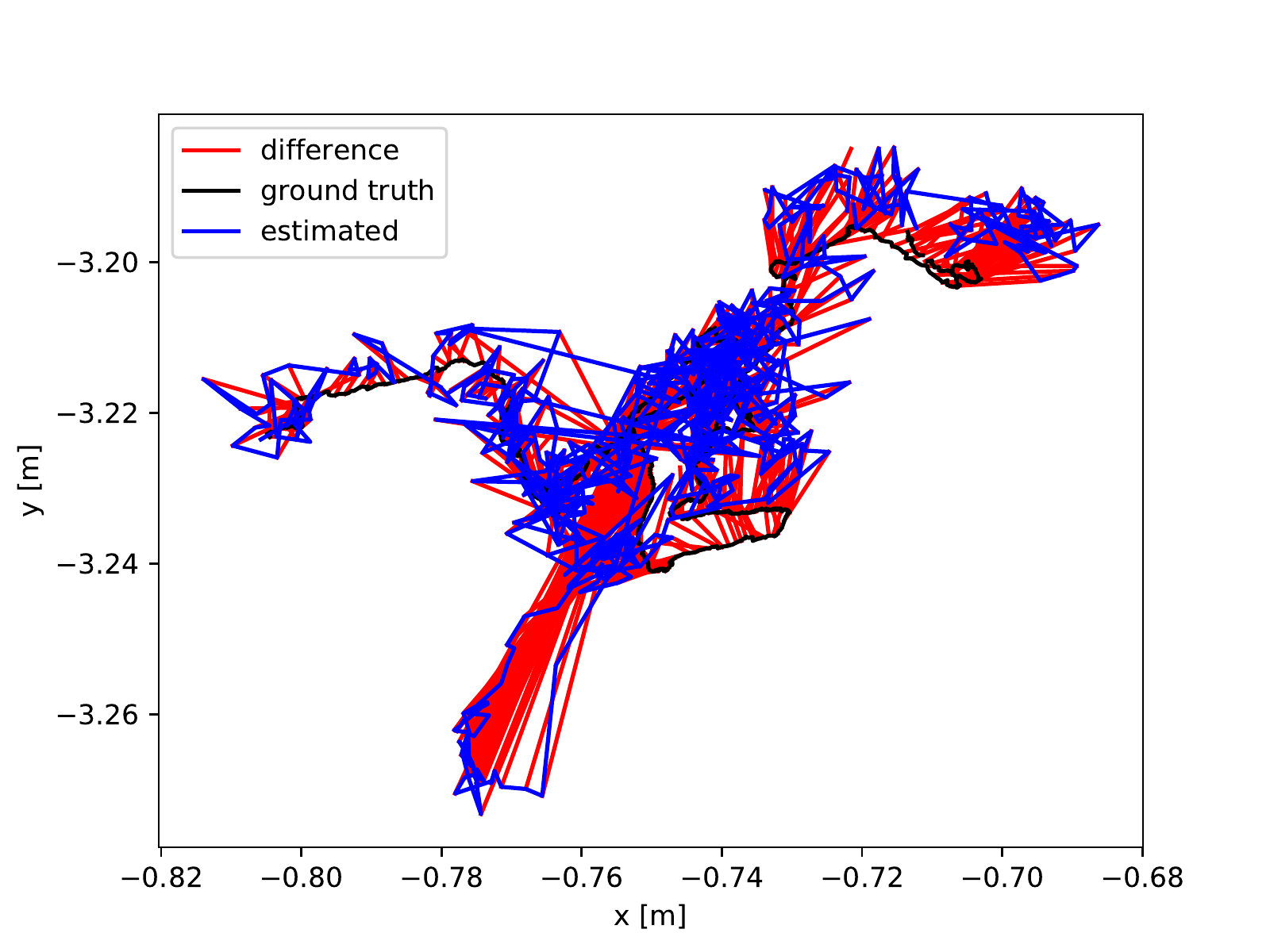}
	\includegraphics[width=0.19\textwidth]{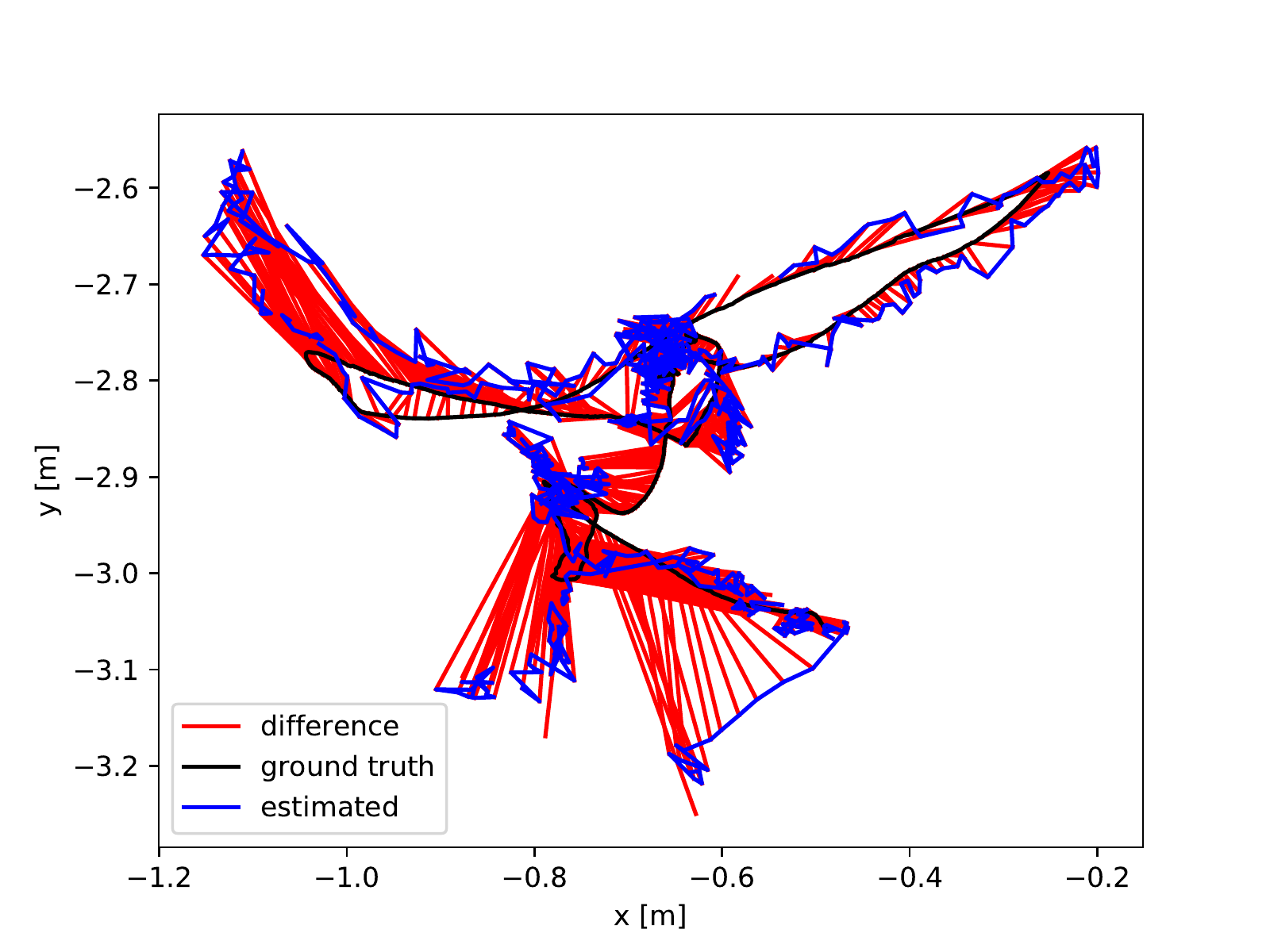}
	\includegraphics[width=0.19\textwidth]{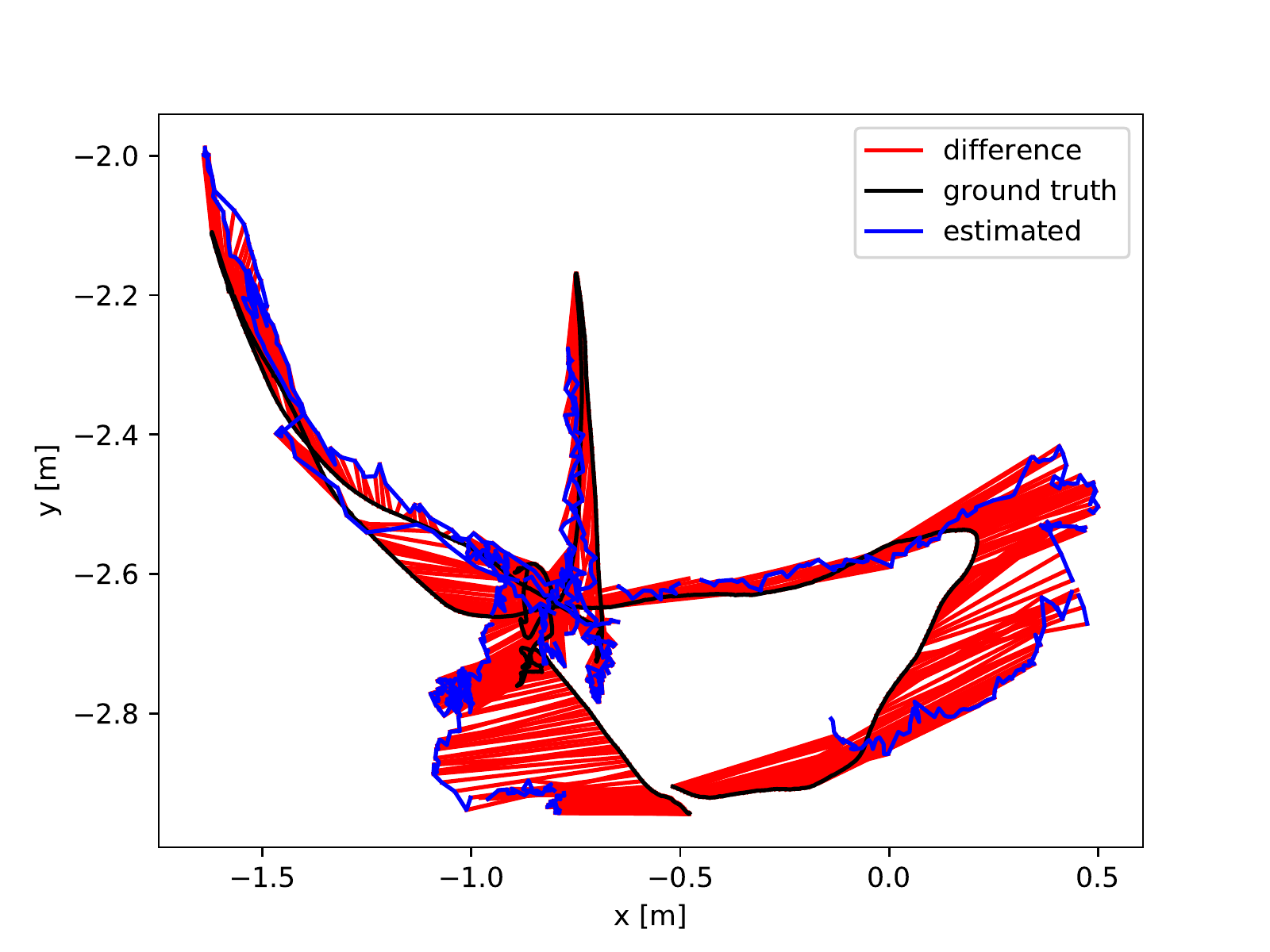}
	\includegraphics[width=0.19\textwidth]{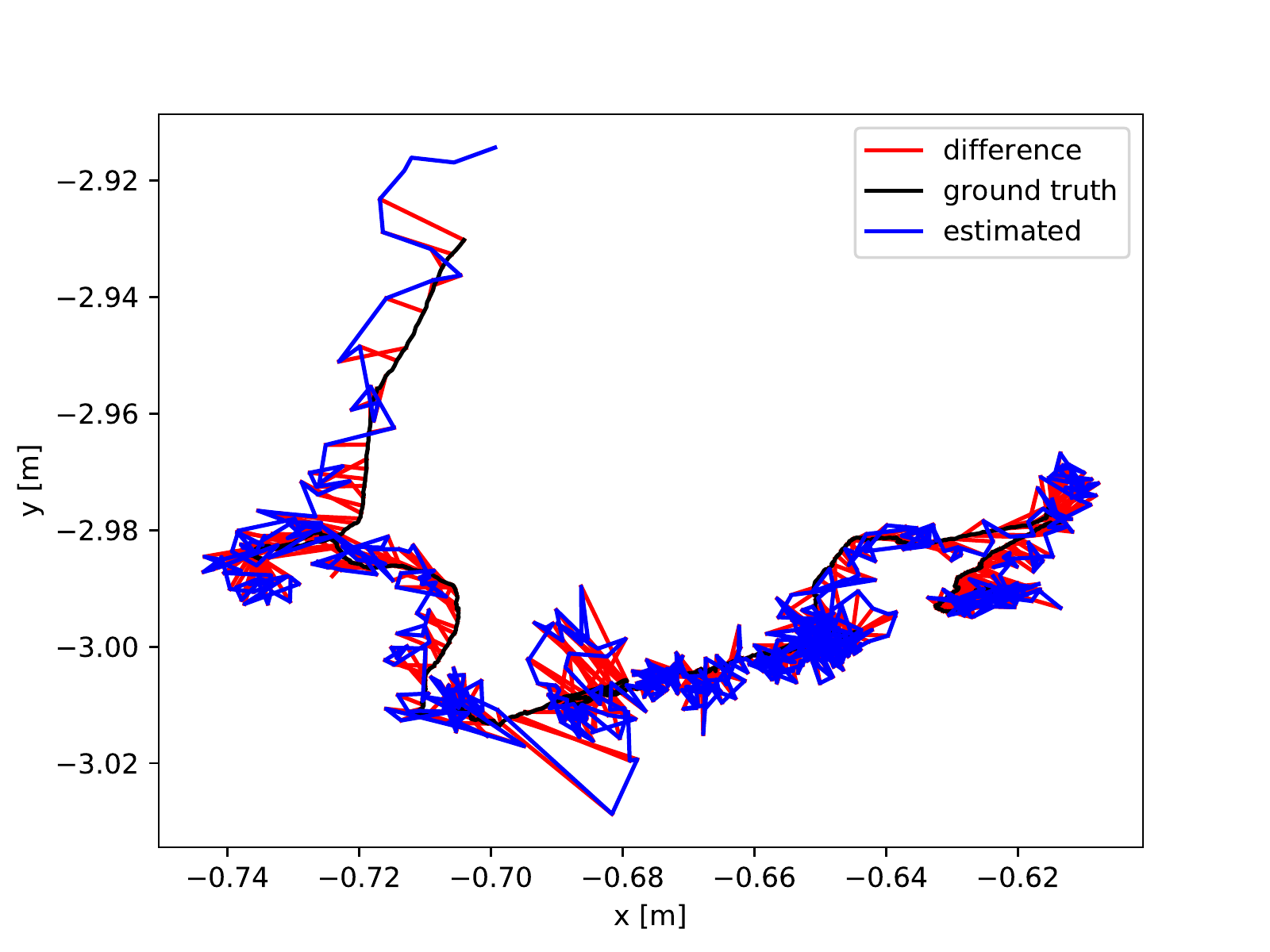} \\
	\includegraphics[width=0.19\textwidth]{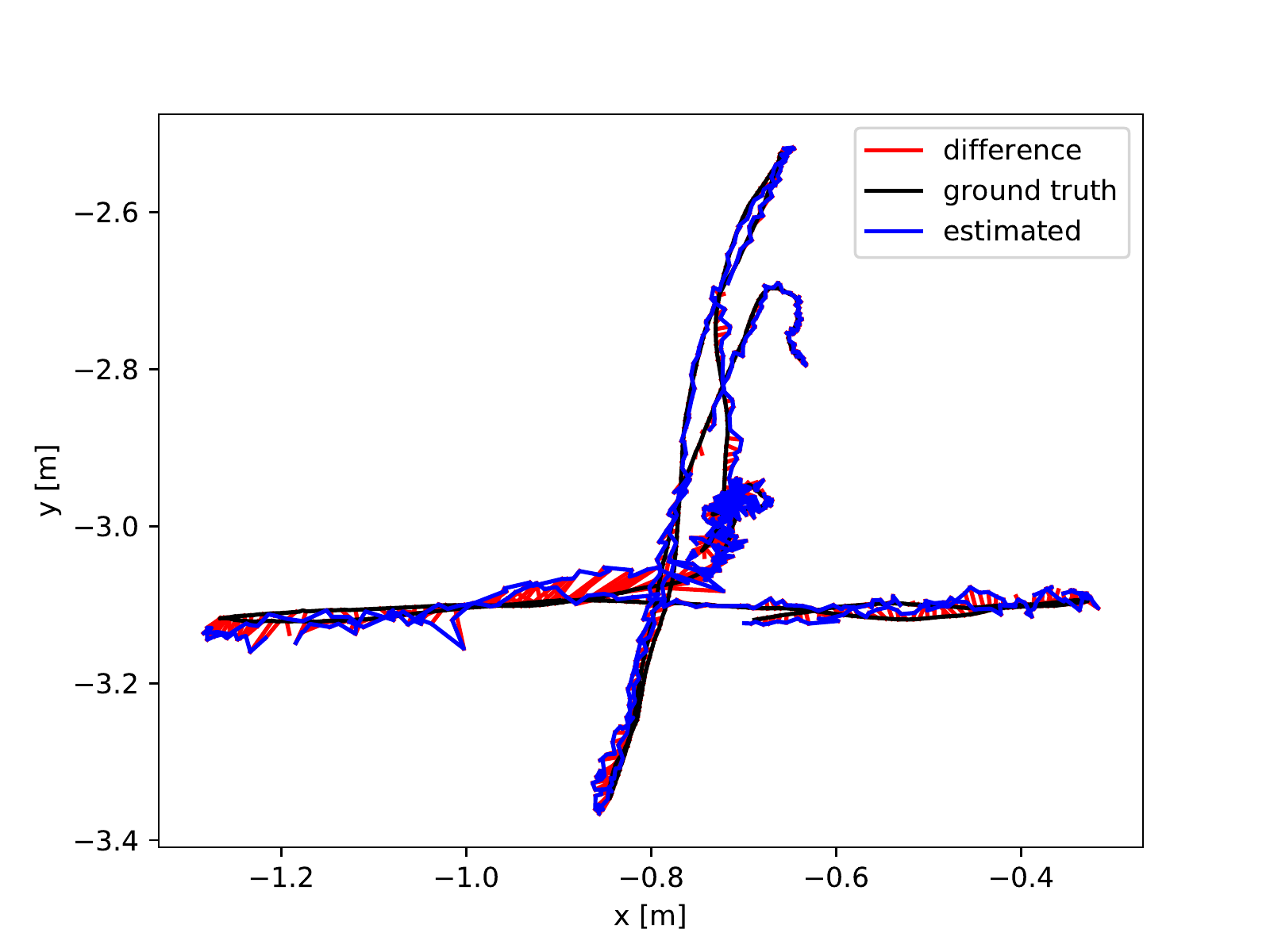}
	\includegraphics[width=0.19\textwidth]{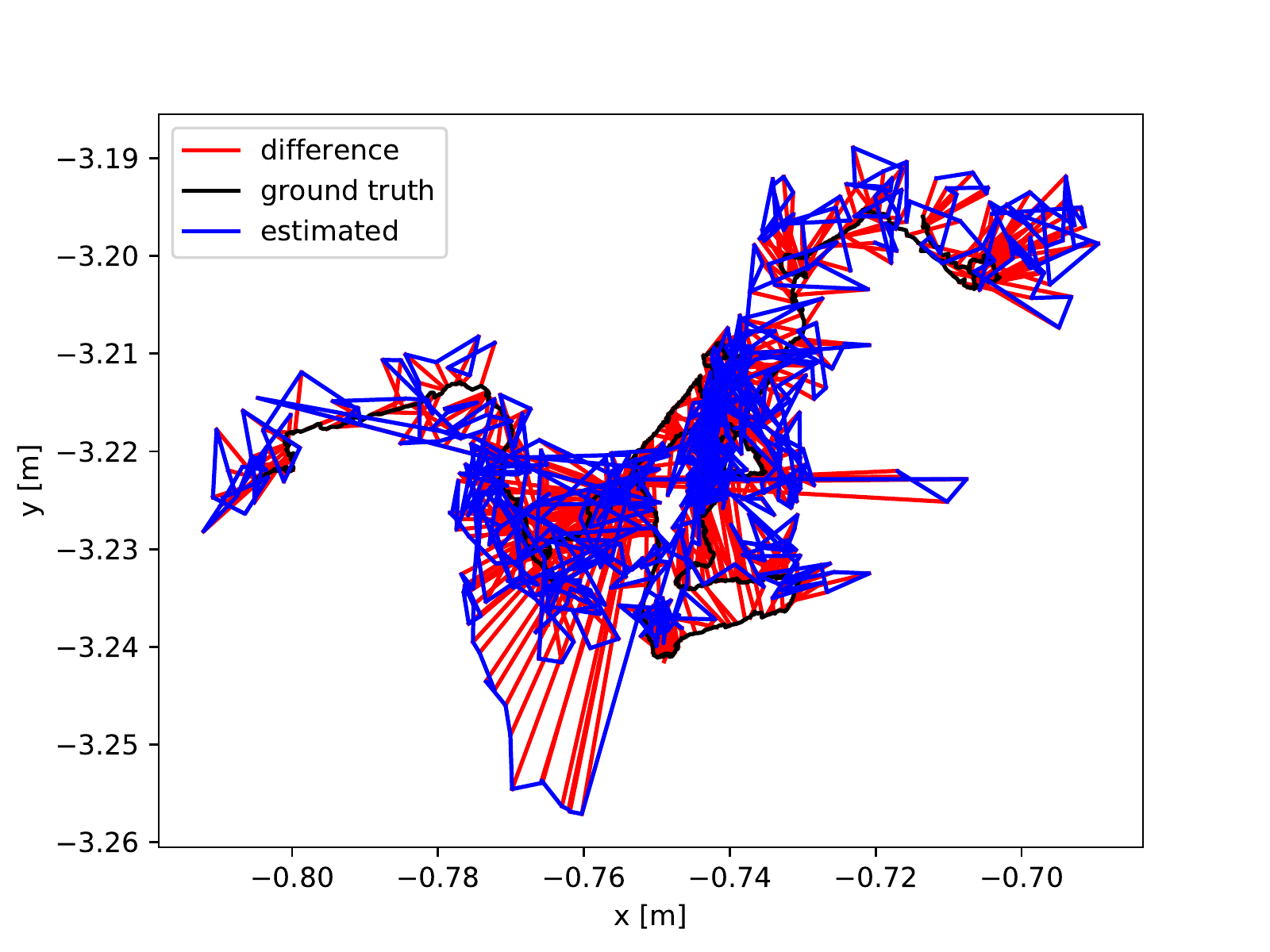}
	\includegraphics[width=0.19\textwidth]{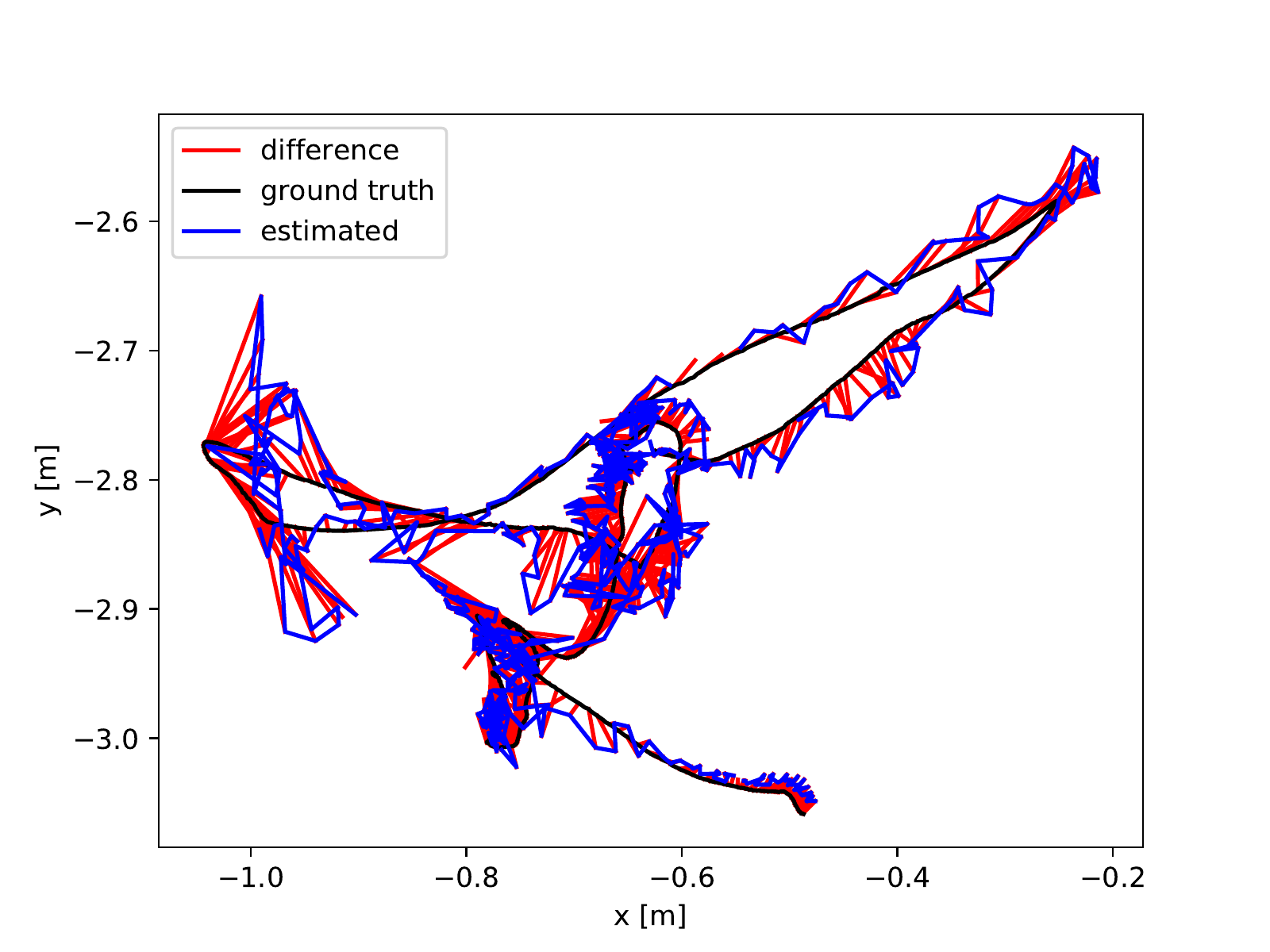}
	\includegraphics[width=0.19\textwidth]{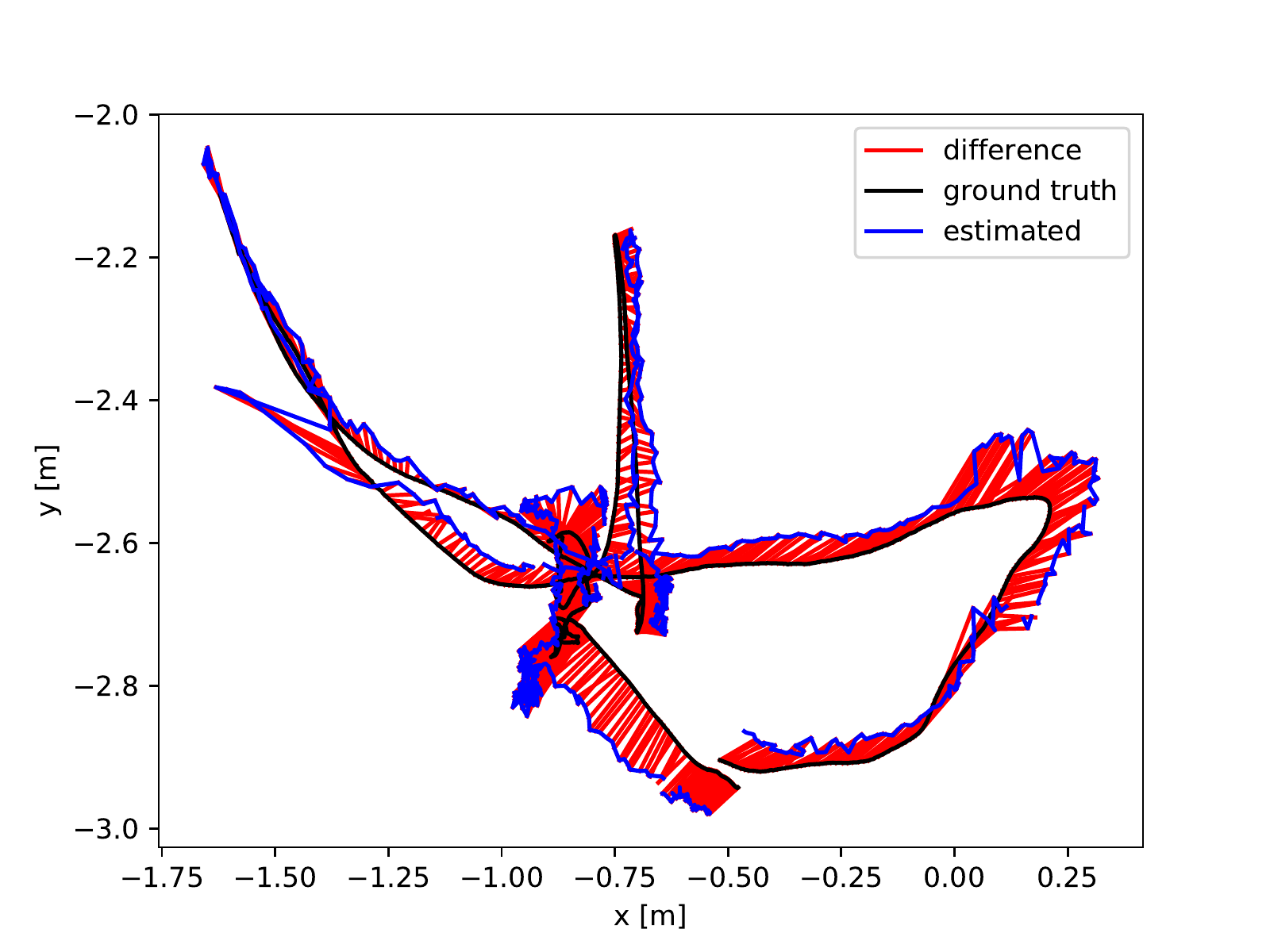}
	\includegraphics[width=0.19\textwidth]{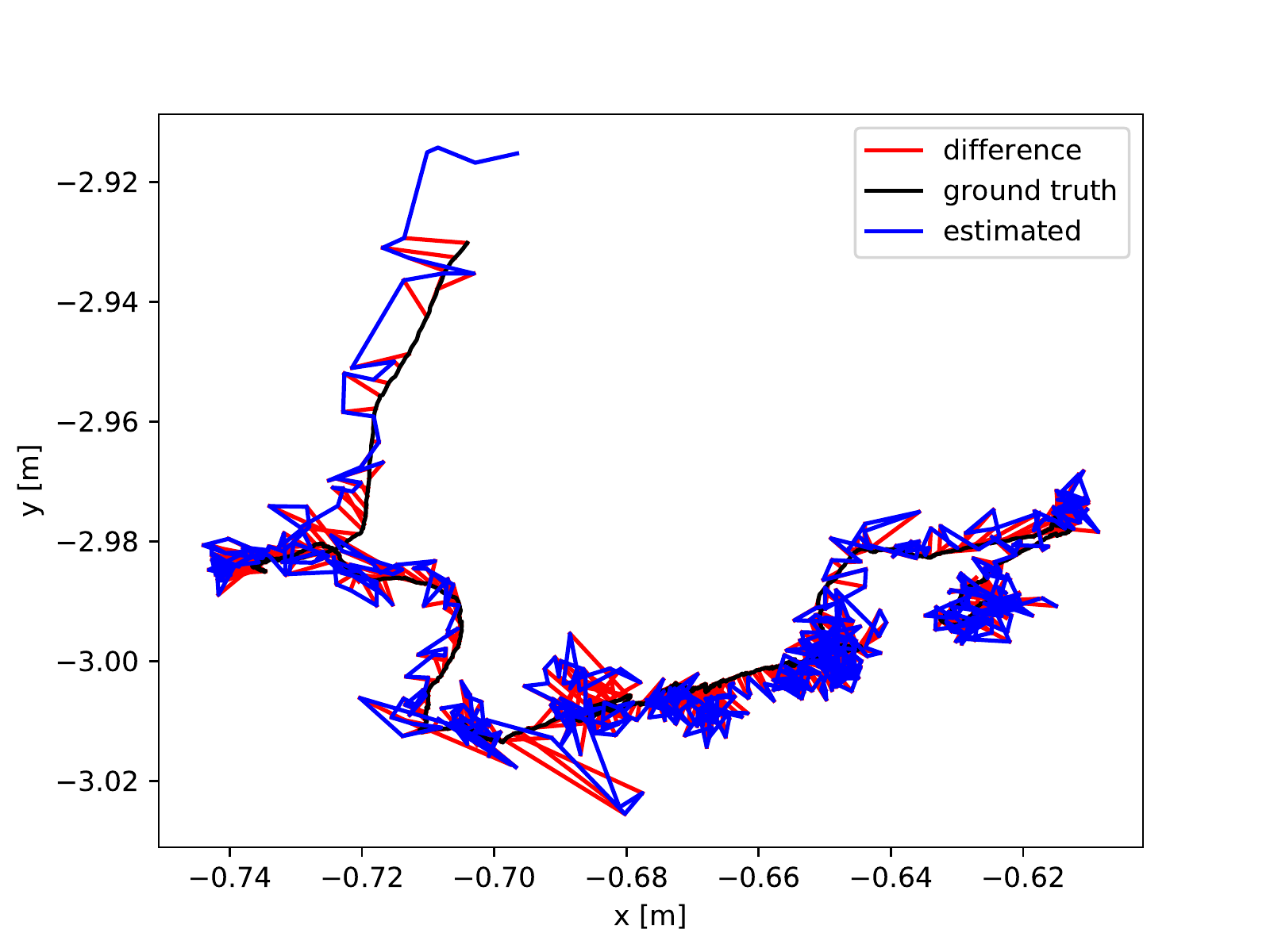} \\
	\includegraphics[width=0.19\textwidth]{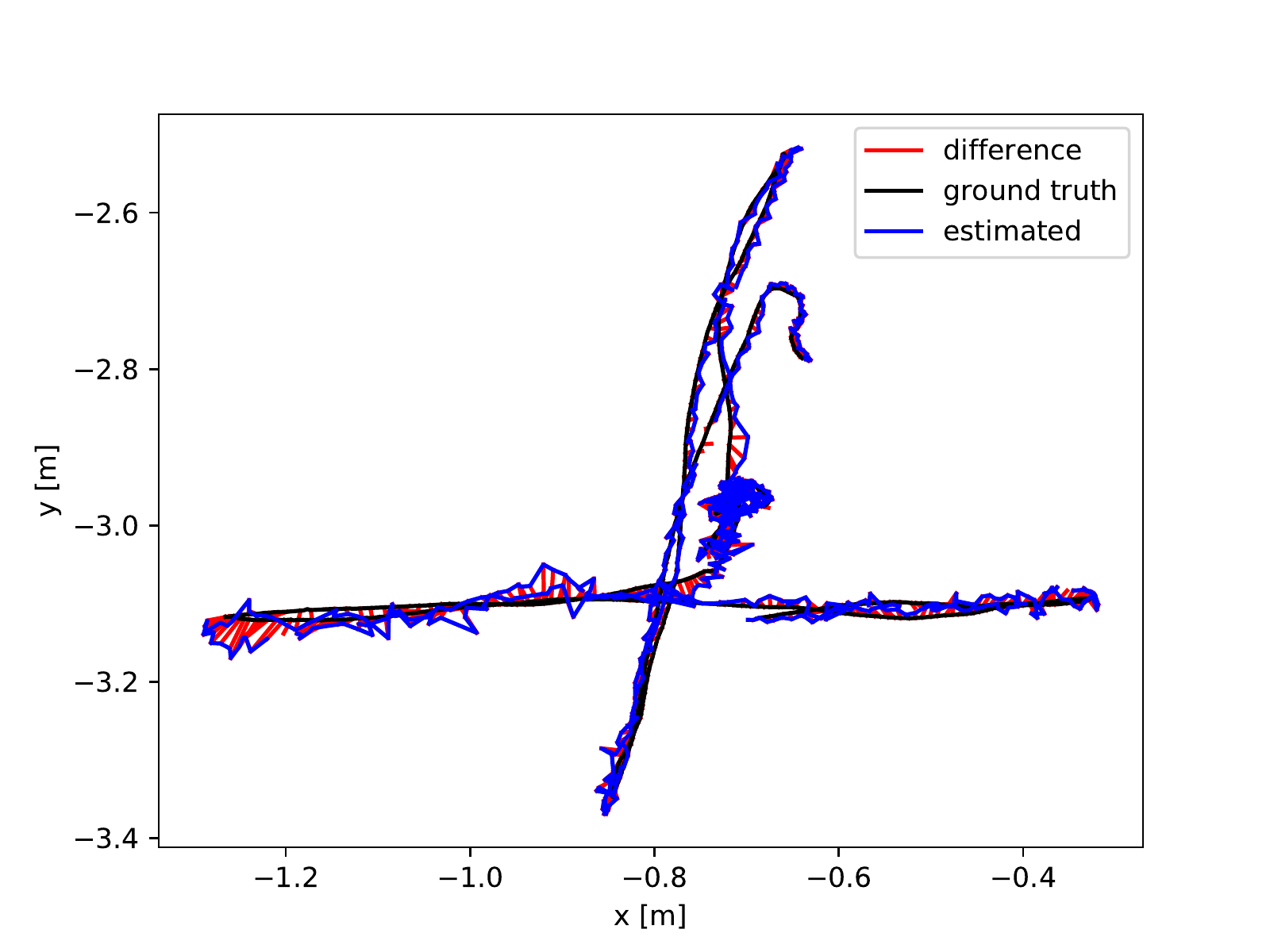}
	\includegraphics[width=0.19\textwidth]{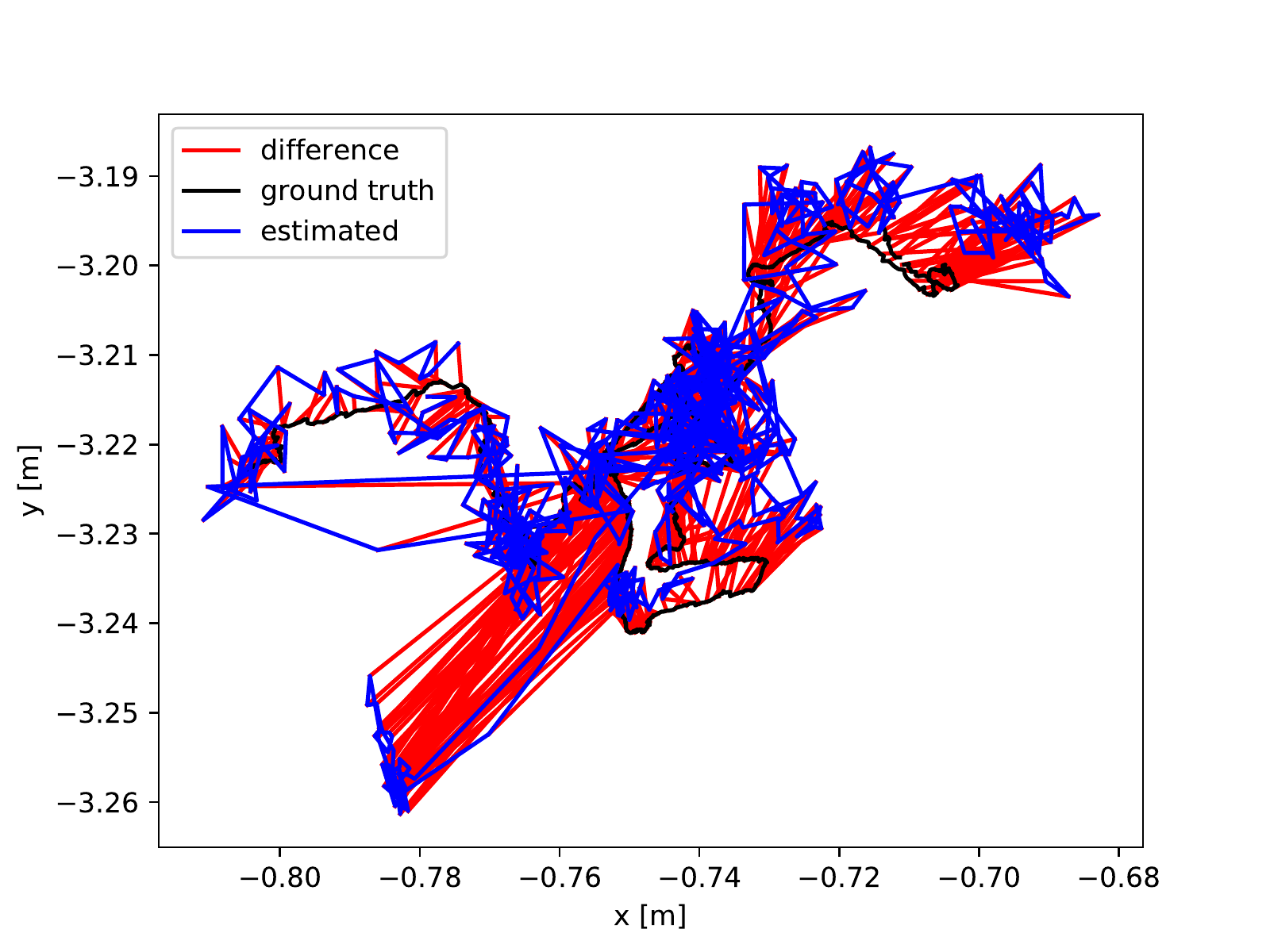}
	\includegraphics[width=0.19\textwidth]{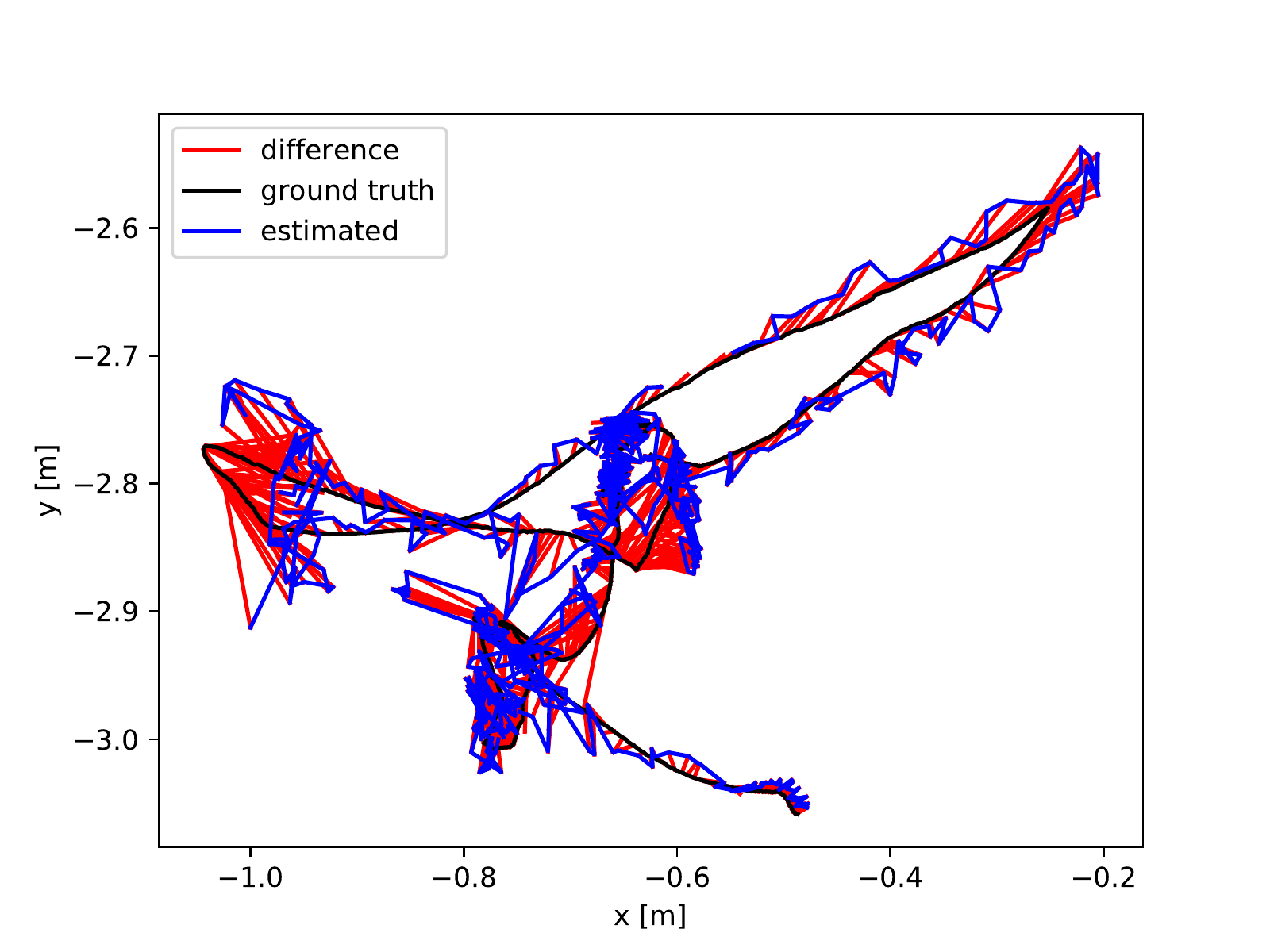}
	\includegraphics[width=0.19\textwidth]{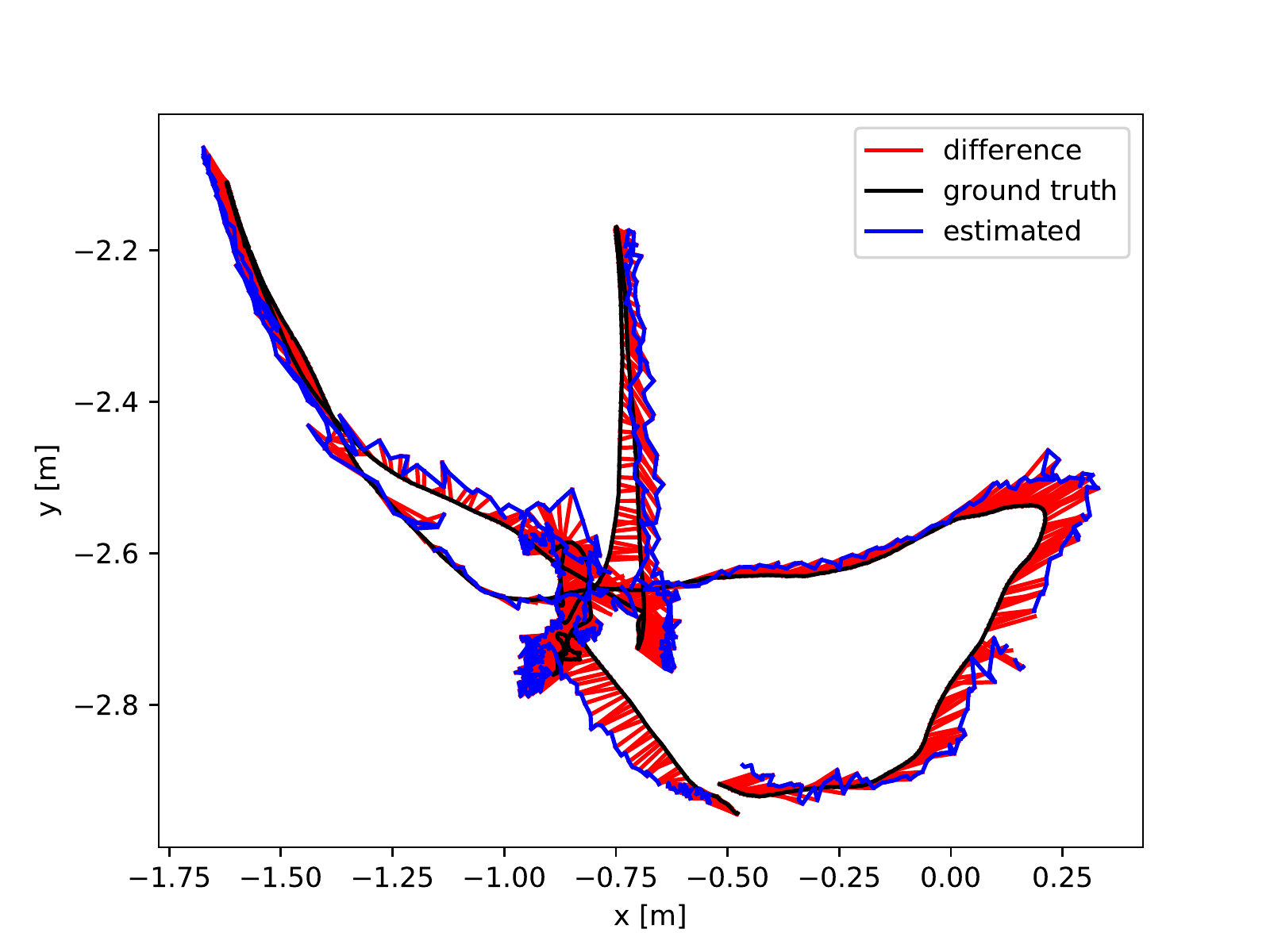}
	\includegraphics[width=0.19\textwidth]{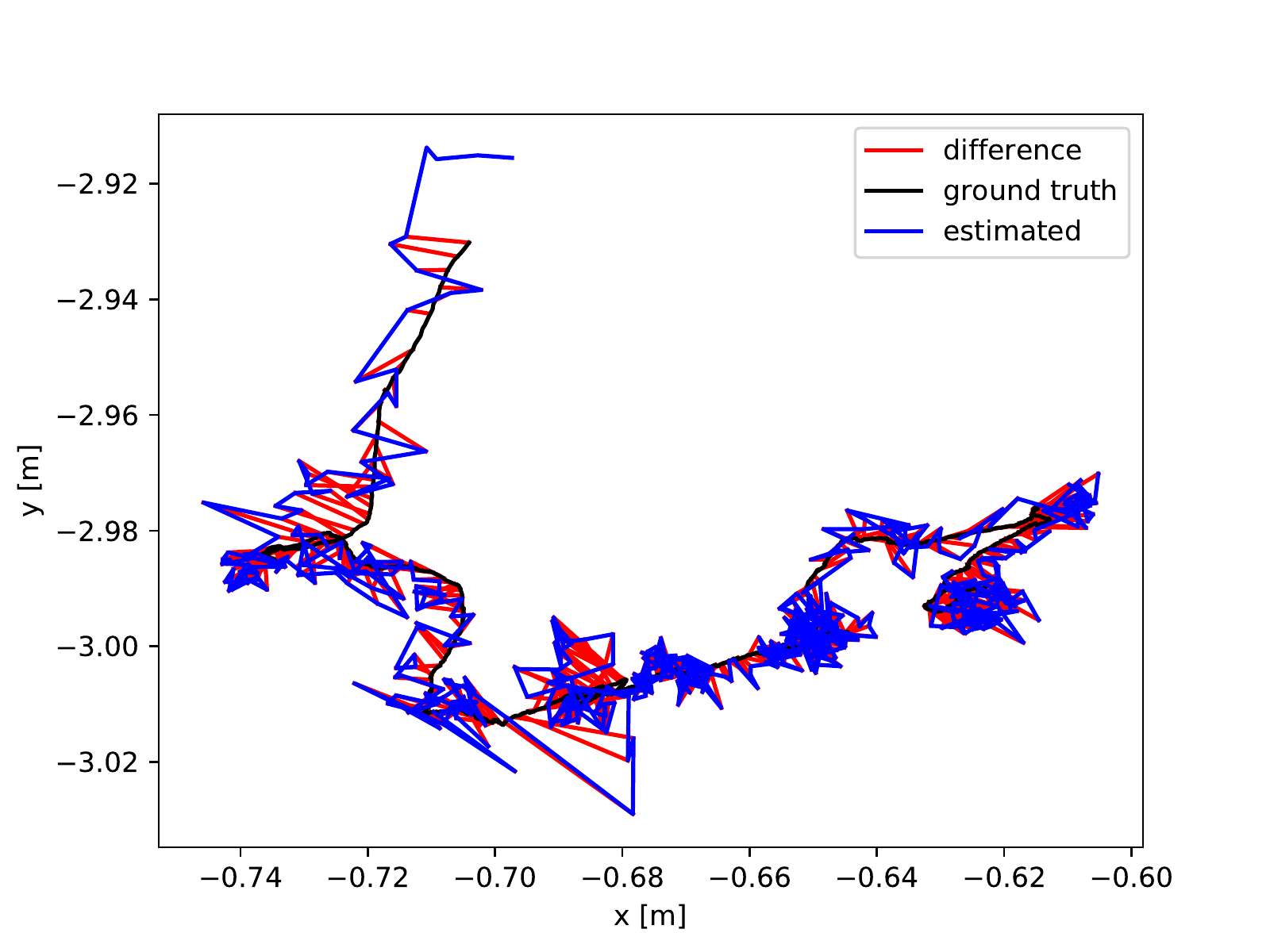} \\
	\caption{Trajectories of sequences (from the left column to the right) \textit{f3/w/xyz},  \textit{f3/w/rpy}, \textit{f3/w/static}, and \textit{f3/w/halfsphere},  \textit{f3/s/static} from dynamic objects sequences of TUM RGB-D dataset \cite{sturm12iros}.  In each column from top to down shows the trajectory generated by: \textit{ORB-SLAM2} (baseline), \textit{+c}, \textit{+d} and \textit{+c,d}.}.
	\label{fig:tum traj}
	\vspace{-7mm}
\end{figure*}

The numerical results of dynamic objects sequences of TUM RGB-D dataset \cite{sturm12iros} and that of its corresponding evaluation sequences, are shown in Table~\ref{tab: tum} and Table~\ref{tab: tum eval} respectively.  The predicted trajectories are shown in Figure~\ref{fig:tum traj} where each row contains the results of the same sequence and each column are of the same method.  The trajectories in the first row are generated by ORB-SLAM2 \cite{murORB2}, which serves as our baseline.  The results of our proposed method with different configurations from the second row to the bottom row are \textit{+c}, \textit{+d}, and \textit{+ c, d}.  

It can be seen that the proposed method significantly outperforms our baseline, i.e. ORB-SLAM2 \cite{murORB2}, especially on the walking sequences which depict high-dynamic scenarios.  In the low-dynamic sequence \textit{f3/s/static}, two sitting people keep static for a long time without big motion, so they can actually be treated as landmarks and offer valid feature points for odometry during a short time but not for long-term data association.  Our method does not rely on temporal static movable objects and reserves the stable number of feature points used in each frame.  It outperforms the ORB-SLAM2 \cite{murORB2} in this low-dynamic sequences, and it has lower chance of erroneous data association during loop closure detection for the long-term concern.

Comparing the results of configurations `+c' `+d' and `+c, d' in Table~\ref{tab: tum}, Table~\ref{tab: tum eval} and the last three rows in Figure~\ref{fig:tum traj}, we found that they are quite similar to each other.  This means using the color image or the depth image or both of them offer similarly effective pairwise potential terms in CRF for TUM RGB-D dataset \cite{sturm12iros}.  We believe `+c,d' will outperform the other two in the situation when the moving objects and the background have similar color pattern or the moving objects are very close to background objects.   

\begin{table*}[!h]
\begin{center}
\setlength\tabcolsep{2.5pt}
\begin{tabular}{ c | c | c c c c c | c c c c c | c c c c c}
\hline
\multirow{2}{5em}{Sequences}  & \multirow{2}{4em}{Methods} & \multicolumn{5}{c}{Absolute Trajectory Error (ATE)} & \multicolumn{5}{c}{Translational Relative Pose Error (RPE)} & \multicolumn{5}{c}{Rotational Relative Pose Error (RPE)} \\ 
\cline{3-17} 
& & RMSE   & Mean   & Median & S.D.   & Improve & RMSE   & Mean   & Median & S.D.   & Improve & RMSE   & Mean   & Median & S.D.   & Improve\\ 
\hline 
\multirow{5}{5em}{f3/w/xyz}  
& ORB-SLAM2 &0.7246&0.6211&0.5562&0.3732&0.00\%&0.4069&0.3014&0.2152&0.2734&0.00\%&7.6817&5.7184&4.0930&5.1292&0.00\%\\
& + c &0.0180&0.0150&0.0128&0.0099&97.52\%&0.0239&0.0194&0.0160&0.0140&94.12\%&\textbf{0.6159}&0.4845&0.4082&0.3802&\textbf{91.98\%}\\
& + d &0.0178&0.0148&0.0127&0.0100&97.54\%&0.0236&0.0196&0.0166&0.0132&94.19\%&0.6368&0.5073&0.4288&0.3849&91.71\%\\
& + c, d &\textbf{0.0176}&0.0153&0.0139&0.0087&\textbf{97.57\%}&\textbf{0.0227}&0.0196&0.0171&0.0114&\textbf{94.43\%}&0.6306&0.5029&0.4204&0.3805&91.79\%\\
\hline
\multirow{5}{5em}{f3/w/static}
& ORB-SLAM2&0.0243&0.0141&0.0103&0.0197&0.00\%&0.0334&0.0182&0.0099&0.0280&0.00\%&0.6020&0.3974&0.2862&0.4523&0.00\%\\
& + c &0.0150&0.0104&0.0075&0.0108&37.97\%&0.0169&0.0122&0.0089&0.0116&49.50\%&\textbf{0.2729}&0.2470&0.2344&0.1161&\textbf{54.67\%}\\
& + d & \textbf{0.0103}&0.0077&0.0059&0.0068&\textbf{57.67\%}&\textbf{0.0139}&0.0107&0.0080&0.0088&\textbf{58.44\%}&0.3051&0.2665&0.2348&0.1484&49.33\%\\
& + c, d &0.0153&0.0108&0.0073&0.0109&36.78\%&0.0165&0.0118&0.0083&0.0115&50.64\%&0.2867&0.2541&0.2350&0.1327&52.38\%\\
\hline
\multirow{5}{5em}{f3/w/rpy}  
& ORB-SLAM2 &1.0326&0.8502&0.8504&0.5861&0.00\%&0.4005&0.2833&0.1470&0.2831&0.00\%&7.7591&5.5058&2.8273&5.4671&0.00\%\\
& + c &0.0959&0.0718&0.0483&0.0635&90.72\%&0.0989&0.0765&0.0535&0.0627&75.30\%&1.8094&1.4542&1.1250&1.0767&76.68\%\\
& + d &\textbf{0.0401}&0.0316&0.0254&0.0247&\textbf{96.11\%}&\textbf{0.0550}&0.0427&0.0330&0.0346&\textbf{86.26\%}&\textbf{1.3152}&0.9897&0.7185&0.8663&\textbf{83.05\%}\\
& + c, d &0.0431&0.0346&0.02637&0.0257&95.83\%&0.0606&0.0485&0.0385&0.0363&84.87\%&1.3728&1.0773&0.8389&0.8509&82.31\%\\
\hline
\multirow{5}{5em}{f3/w/
	halfsphere}
& ORB-SLAM2 &0.4295&0.3895&0.3239&0.1810&0.00\%&0.2773&0.1743&0.0642&0.2156&0.00\%&5.1820&3.3064&1.5104&3.9901&0.00\%\\
& + c & 0.2273&0.1987&0.1674&0.1105&47.07\%&0.1046&0.0696&0.0438&0.0781&62.28\%&2.2816&1.5433&1.0037&1.6804&55.97\%\\
& + d &0.0873&0.0823&0.0816&0.0291&79.67\%&0.0549&0.0396&0.0293&0.0381&80.19\%&1.1941&0.8743&0.7037&0.8134&76.96\%\\
& + c, d &\textbf{0.0774}&0.0744&0.0729&0.0213&\textbf{81.98\%}&\textbf{0.0482}&0.0363&0.0268&0.0317&\textbf{82.62\%}&\textbf{1.0390}&0.8520&0.7292&0.5946&\textbf{79.95\%}\\
\hline
\multirow{5}{5em}{f3/s/static}
& ORB-SLAM2 &0.0089&0.0078&0.0069&0.0043&0.00\%&0.0098&0.0087&0.0078&0.0046&0.00\%&0.2902&0.2628&0.2513&0.1231&0.00\%\\
& + c &0.0063&0.0055&0.0049&0.0030&29.30\%&0.0078&0.0068&0.0062&0.0038&20.74\%&\textbf{0.2640}&0.2358&0.2208&0.1186&\textbf{9.05\%}\\
& + d &\textbf{0.0058}&0.0050&0.0043&0.0029&\textbf{34.89\%}&\textbf{0.0071}&0.0063&0.0057&0.0032&\textbf{28.16\%}&0.2674&0.2414&0.2318&0.1150&7.86\%\\
& + c, d &0.0062&0.0054&0.0047&0.0032&30.16\%&0.0074&0.0065&0.0058&0.0035&24.75\%&0.2673&0.2408&0.2286&0.1160&7.90\%\\
\hline
\end{tabular}
\end{center}
\caption{Experimental results on dynamic objects sequences of TUM RGB-D dataset \cite{sturm12iros}.  The best performed methods are highlighted by boldface.}
\label{tab: tum}
\vspace{-7mm}
\end{table*}

\begin{table*}[!h]
\begin{center}
\setlength\tabcolsep{2.5pt}
\begin{tabular}{ c | c | c c c c c | c c c c c | c c c c c}
\hline
\multirow{2}{5em}{Sequences}  & \multirow{2}{4em}{Methods} & \multicolumn{5}{c}{Absolute Trajectory Error (ATE)} & \multicolumn{5}{c}{Translational Relative Pose Error (RPE)} & \multicolumn{5}{c}{Rotational Relative Pose Error (RPE)} \\ 
\cline{3-17} 
 & & RMSE   & Mean   & Median & S.D.   & Improve & RMSE   & Mean   & Median & S.D.   & Improve & RMSE   & Mean   & Median & S.D.   & Improve\\ 
\hline 
\multirow{5}{5em}{f3/w/xyz/v}  
& ORB-SLAM2 & 1.0301 & 0.8908 & 0.7737 & 0.5172 &0.00\%&0.3271&0.1402&0.0215&0.2956&0.00\%&6.0940&2.7293&0.5963&5.4487&0.00\%  \\  
& + c & \textbf{0.0115}&0.0103&0.0094&0.0052&\textbf{98.88\%}&\textbf{0.0151}&0.0133&0.0120&0.0071&\textbf{95.39\%}&\textbf{0.4861}&0.4369&0.4123&0.2132&\textbf{92.02\%}\\
& + d &0.0126&0.0112&0.0099&0.0059&98.77\%&0.0166&0.0145&0.0129&0.0080&94.93\%&0.4978&0.44607&0.4120&0.2210&91.83\%\\
& + c, d &0.0120&0.0107&0.0097&0.0053&98.84\%&0.0157&0.0140&0.0125&0.0073&95.19\%&0.4921&0.4420&0.4159&0.2164&91.92\%\\
\hline
\multirow{5}{5em}{f3/w/static/v}
& ORB-SLAM2 &0.9144&0.8898&1.0002&0.2107&0.00\%&0.2748&0.1459&0.0502&0.2329&0.00\%&5.0884&2.7055&0.9576&4.3095&0.00\% \\
& + c &\textbf{0.0084}&0.0071&0.0063&0.0045&\textbf{99.08\%}&\textbf{0.0111}&0.0094&0.0084&0.0060&\textbf{95.94\%}&\textbf{0.3033}&0.2677&0.2463&0.1426&\textbf{94.04\%}\\
& + d &0.0096&0.0078&0.0066&0.0056&98.95\%&0.0120&0.0099&0.0084&0.0068&95.63\%&0.3192&0.2764&0.2442&0.1595&93.73\%\\
& + c, d &0.0094&0.0082&0.0076&0.0045&98.98\%&0.0114&0.0099&0.0087&0.0057&95.84\%&0.3051&0.2731&0.2509&0.1361&94.00\%\\
\hline
\multirow{5}{5em}{f3/w/rpy/v}  
& ORB-SLAM2&0.6108&0.4512&0.2910&0.4117&0.00\%&0.3505&0.1859&0.0470&0.2972&0.00\%&6.8413&3.7676&1.2285&5.7104&0.00\% \\ 
& + c &0.0297&0.0245&0.0191&0.0168&95.13\%&\textbf{0.0355}&0.0293&0.0236&0.0200&\textbf{89.88\%}&\textbf{0.8981}&0.7646&0.6556&0.4712&\textbf{86.87\%}\\
& + d &0.0293&0.0240&0.0191&0.0168&95.20\%&0.0365&0.0298&0.0237&0.0211&89.57\%&0.9484&0.7958&0.6743&0.5159&86.14\%\\
& + c, d &\textbf{0.0291}&0.0235&0.0185&0.0171&\textbf{95.24\%}&0.0367&0.0287&0.0223&0.0228&89.53\%&0.9437&0.7728&0.6585&0.5416&86.21\%\\
\hline
\multirow{5}{5em}{f3/w/
	halfsphere/v}
& ORB-SLAM2&0.5903&0.4924&0.5020&0.3256&0.00\%&0.2838&0.1763&0.0766&0.2224&0.00\%&5.5842&3.5018&1.6032&4.3498&0.00\%\\
& + c &0.4313&0.3563&0.3246&0.2430&26.94\%&0.2094&0.1227&0.0502&0.1697&26.23\%&4.8268&2.7929&1.0360&3.9367&13.56\%\\
& + d &\textbf{0.0470}&0.0392&0.0308&0.0259&\textbf{92.03\%}&\textbf{0.0411}&0.0329&0.0239&0.0247&\textbf{85.51\%}&\textbf{0.9303}&0.7693&0.6198&0.5230&\textbf{83.34\%}\\
& + c, d &0.1972&0.1387&0.0835&0.1401&66.60\%&0.1241&0.0649&0.0301&0.1058&56.29\%&2.6487&1.3642&0.7151&2.2703&52.57\%\\
\hline
\multirow{5}{5em}{f3/s/static/v}
& ORB-SLAM2&0.0056&0.0051&0.0047&0.0023&0.00\%&\textbf{0.0066}&0.0057&0.0052&0.0032&\textbf{0.00\%}&\textbf{0.3025}&0.2668&0.2359&0.1424&\textbf{0.00\%}\\
& + c &0.0056&0.0048&0.0043&0.0028&0.87\%&0.0068&0.0060&0.0053&0.0033&-3.67\%&0.3080&0.2726&0.2484&0.1432&-1.82\%\\
& + d &\textbf{0.0054}&0.0047&0.0042&0.0027&\textbf{3.79\%}&0.0070&0.0061&0.0056&0.0034&-5.75\%&0.3096&0.2740&0.2503&0.1442&-2.37\%\\
& + c, d &0.0056&0.0049&0.0042&0.0027&0.87\%&0.0070&0.0062&0.0055&0.0033&-6.47\%&0.3105&0.2745&0.2415&0.1451&-2.67\%\\
\hline
\end{tabular}
\end{center}
\caption{Experimental results on dynamic objects validation sequences of TUM RGB-D dataset \cite{sturm12iros}.  The best performed methods are highlighted by boldface.}
\label{tab: tum eval}
\vspace{-7mm}
\end{table*}

Our method is compared with others by the improvement of RMSE in equation (\ref{equ:improvement}).  Since different methods have different baselines, different implementation, different metric units, and there are randomness in each run, etc.  We think this comparison is relatively fair.  We pick most recent methods that are modified from either ORB-SLAM2 \cite{murORB2} or DVO \cite{kerl2013dense}, and test on TUM RGB-D dataset \cite{sturm12iros}. The comparison results are shown in Table~\ref{tab: tum compare}, where RPE.T and RPE.R mean the translational part and rotational part of RPE.  The methods based on DVO \cite{kerl2013dense} are noted with superscript, and the rest methods are based on ORB-SLAM2 \cite{murORB2}.  The best results are highlighted by boldface.  It can be seen that the proposed method achieve the state-of-the-art performance on 53.3\% of the results, and are comparable with other methods for the rest.  Even though our semantic segmentation module is weakly supervised, our overall system outperforms DS-SLAM \cite{yu2018ds}, which adopts the fully-supervised SegNet \cite{badrinarayanan2017segnet}.  

\begin{table}[!h]
\begin{center}
\begin{tabular}{ c | c | c c c }
\hline	
\multirow{2}{4em}{Sequences}  & \multirow{2}{4em}{Methods} & \multicolumn{3}{c}{RMSE Improvement} \\	
\cline{3-5}
 & & ATE   & RPE.T & RPE.R \\
\hline
\multirow{7}{4em}{f3/w/xyz}
& DS-SLAM \cite{yu2018ds} & 96.71\% &91.93\% &89.32\% \\
& Unified \cite{wang2018unified} & 96.73\% & -- & -- \\
& DynaSLAM \cite{bescos2018dynaslam} & 96.73\% & -- & -- \\
& Detect-SLAM \cite{f2018detect} & \textbf{97.62\%} & -- & -- \\
& Improving\textsuperscript{DVO} \cite{sun2017improving} & 84.38\% & 69.78\% & 62.65\% \\
& M-removal\textsuperscript{DVO} \cite{sun2018motion} & 88.99\% & 83.37\% & 81.58\% \\
& \textbf{Ours} & 97.57\% & \textbf{94.43\%} & \textbf{91.79\%} \\
\hline
\multirow{7}{4em}{f3/w/static}
& DS-SLAM \cite{yu2018ds} & \textbf{97.91\%} & \textbf{95.27\%} & \textbf{93.09\%} \\
& Unified \cite{wang2018unified} & 93.33\% & -- & -- \\
& DynaSLAM \cite{bescos2018dynaslam} & 93.33\% & -- & -- \\
& Detect-SLAM \cite{f2018detect} & -- & -- & -- \\
& Improving\textsuperscript{DVO} \cite{sun2017improving} & 69.06\% & 65.65\% & 52.09\% \\
& M-removal\textsuperscript{DVO} \cite{sun2018motion} & 84.25\% & 87.47\% & 78.96\% \\
& \textbf{Ours} & 36.78\% & 50.64\% & 52.38\% \\
\hline
\multirow{7}{4em}{f3/w/rpy}
& DS-SLAM \cite{yu2018ds} & 48.97\% & 64.64\% & 62.82\% \\
& Unified \cite{wang2018unified} & 94.71\% & -- & -- \\
& DynaSLAM \cite{bescos2018dynaslam} & 94.71\% & -- & -- \\
& Detect-SLAM \cite{f2018detect} & 66.94\% & -- & -- \\
& Improving\textsuperscript{DVO} \cite{sun2017improving} & 81.75\% & 62.30\% & 51.54\% \\
& M-removal\textsuperscript{DVO} \cite{sun2018motion} & 90.02\% & 79.15\% & 71.28\% \\
& \textbf{Ours} & \textbf{95.83\%} & \textbf{84.87\%} & \textbf{82.31\%} \\
\hline
\multirow{7}{4em}{f3/w/
	halfsphere}
& DS-SLAM \cite{yu2018ds} & \textbf{93.76\%} & \textbf{91.62\%} & \textbf{88.96\%} \\
& Unified \cite{wang2018unified} & 92.88\% & -- & -- \\
& DynaSLAM \cite{bescos2018dynaslam} & 92.88\% & -- & -- \\
& Detect-SLAM \cite{f2018detect} & 91.18\% & -- & -- \\
& Improving\textsuperscript{DVO} \cite{sun2017improving} & 76.32\% & 49.09\% & 24.22\% \\
& M-removal\textsuperscript{DVO} \cite{sun2018motion} & 87.37\% & 81.38\% & 71.26\% \\
& \textbf{Ours} & 81.98\% & 82.62\% & 79.95\% \\
\hline
\multirow{7}{4em}{f3/s/static}
& DS-SLAM \cite{yu2018ds} & 25.94\% & 17.61\% & 5.07\% \\
& Unified \cite{wang2018unified} & -66.67\% & -- & -- \\
& DynaSLAM \cite{bescos2018dynaslam} & -- & -- & -- \\
& Detect-SLAM \cite{f2018detect} & -- & -- & -- \\
& Improving\textsuperscript{DVO} \cite{sun2017improving} & -- & -- & -- \\
& M-removal\textsuperscript{DVO} \cite{sun2018motion} & -- & -- & -- \\
& \textbf{Ours} & \textbf{30.16\%} & \textbf{24.75\%} & \textbf{7.90\%} \\
\hline
\end{tabular}
\end{center}
\caption{Comparison of the improvement of RMSE on dynamic objects sequences of TUM RGB-D dataset \cite{sturm12iros}.  RPE.T and RPE.R mean the translational part and rotational part of RPE.  The methods based on DVO \cite{kerl2013dense} are noted with superscript, and the rest methods are based on ORB-SLAM2 \cite{murORB2}.  The best results are highlighted by boldface.}
\label{tab: tum compare}
\vspace{-7mm}
\end{table}

\subsubsection{Results on stereo KITTI Visual Odometry dataset \cite{kitti}}
The numerical results on stereo KITTI Visual Odometry dataset \cite{kitti} are shown in Table~\ref{tab: kitti}.  It can be seen that the results of our proposed method outperform ORB-SLAM2 \cite{murORB2} in 6 out of 10 sequences.   Significant improvements are shown in sequence 06, 07 and 09, for the rest sequences our method is closely comparable with ORB-SLAM2 \cite{murORB2}.  This is due to the properties of KITTI Visual Odometry dataset \cite{kitti}.  Its sequences are typically captured by driving around the mid-size city of Karlsruhe, and each sequence last for a few minutes.  There are many cars parked at both sides of the road.  These cars are static at the moment when the videos are captured.  The proposed method prevents the feature points falling into the movable objects region, this reduces the chance of erroneous data association for the long-term, but may hinder the short-term performance a little bit.  Overall, our proposed method is much more robust in dynamic environments than our baseline.

\begin{table*}[!h]
\begin{center}[!h]
\begin{tabular}{ c | c | c c c c c | c c c c c }
\hline
\multirow{2}{4em}{Sequences}  & \multirow{2}{4em}{Methods} & \multicolumn{5}{c}{Absolute Trajectory Error (ATE)} & \multicolumn{5}{c}{Translational Relative Pose Error (RPE): meter/frame} \\ 
\cline{3-12} 
& & RMSE   & Mean   & Median & S.D.   & Improve & RMSE   & Mean   & Median & S.D.   & Improve \\ 
\hline 
\multirow{3}{4em}{00}
&ORB-SLAM2&\textbf{1.3196}&\textbf{1.1887}&\textbf{1.0730}&\textbf{0.5731}&\textbf{0.00\%}&\textbf{0.0283}&\textbf{0.0197}&\textbf{0.0152}&\textbf{0.0203}&\textbf{0.00\%}\\
& + c &1.3635&1.2125&1.0939&0.6236&-3.32\%&0.0288&0.0202&0.0155&0.0206&-1.85\%\\
\hline 
\multirow{3}{4em}{01}
&ORB-SLAM2&\textbf{9.6003}&\textbf{8.9647}&\textbf{7.8879}&\textbf{3.4352}&\textbf{0.00\%}&\textbf{0.0509}&\textbf{0.0469}&0.0450&\textbf{0.0196}&\textbf{0.00\%}\\
& + c &11.5804&10.9703&9.9151&3.7093&-20.63\%&0.0513&0.0474&\textbf{0.0449}&0.0198&-0.92\%\\
\hline 
\multirow{3}{4em}{02}
&ORB-SLAM2&6.5433&5.5172&4.3230&3.5179&0.00\%&\textbf{0.0283}&\textbf{0.0230}&\textbf{0.0191}&\textbf{0.0165}&\textbf{0.00\%}\\
& + c &\textbf{5.9395}&\textbf{5.0270}&\textbf{4.1188}&\textbf{3.1633}&\textbf{9.23\%}&0.0288&0.0231&0.0192&0.0171&-1.53\%\\
\hline 
\multirow{3}{4em}{03}
&ORB-SLAM2&\textbf{0.8276}&\textbf{0.7034}&\textbf{0.5378}&\textbf{0.4360}&\textbf{0.00\%}&\textbf{0.0178}&\textbf{0.0156}&\textbf{0.0137}&\textbf{0.0085}&\textbf{0.00\%}\\
& + c &0.9207&0.7943&0.6321&0.4657&-11.25\%&0.0180&0.0157&0.0140&0.0088&-1.00\%\\
\hline 
\multirow{3}{4em}{04}
&ORB-SLAM2&\textbf{0.2319}&\textbf{0.2092}&0.1967&\textbf{0.1002}&\textbf{0.00\%}&0.0221&0.0197&0.0176&0.0101&0.00\%\\
& + c &0.2602&0.2287&\textbf{0.1879}&0.1241&-12.19\%&\textbf{0.0207}&\textbf{0.0185}&\textbf{0.0166}&\textbf{0.0094}&\textbf{6.41\%}\\
\hline 
\multirow{3}{4em}{05}
&ORB-SLAM2&0.7967&0.7100&\textbf{0.6579}&0.3616&0.00\%&0.0169&0.0135&\textbf{0.0113}&0.0102&0.00\%\\
& + c &\textbf{0.7516}&\textbf{0.6862}&0.6736&\textbf{0.3066}&\textbf{5.67\%}&\textbf{0.0166}&\textbf{0.0134}&0.0114&\textbf{0.0097}&\textbf{2.14\%}\\
\hline 
\multirow{3}{4em}{06}
&ORB-SLAM2&0.9557&0.9022&0.9070&0.3151&0.00\%&0.0296&0.0162&0.0122&0.0248&0.00\%\\
& + c &\textbf{0.7984}&\textbf{0.7654}&\textbf{0.8315}&\textbf{0.2271}&\textbf{16.46\%}&\textbf{0.0204}&\textbf{0.0143}&\textbf{0.0115}&\textbf{0.0146}&\textbf{31.01\%}\\
\hline 
\multirow{3}{4em}{07}
&ORB-SLAM2&0.5577&0.5256&0.5595&0.1866&0.00\%&\textbf{0.0168}&\textbf{0.0135}&\textbf{0.0113}&0.0100&\textbf{0.00\%}\\
& + c &\textbf{0.4895}&\textbf{0.4582}&\textbf{0.4263}&\textbf{0.1722}&\textbf{12.23\%}&0.0173&0.0143&0.0120&\textbf{0.0097}&-2.68\%\\
\hline 
\multirow{3}{4em}{08}
&ORB-SLAM2&\textbf{3.5123}&\textbf{3.1886}&2.8263&\textbf{1.4728}&\textbf{0.00\%}&0.0387&0.0247&\textbf{0.0172}&0.0298&0.00\%\\
& + c &3.5694&3.2154&\textbf{2.6988}&1.5498&-1.63\%&\textbf{0.0387}&0.0247&0.0173&\textbf{0.0297}&\textbf{0.04\%}\\
\hline 
\multirow{3}{4em}{09}
&ORB-SLAM2&3.1299&2.6469&2.1294&1.6703&0.00\%&\textbf{0.0220}&\textbf{0.0189}&\textbf{0.0157}&\textbf{0.0112}&\textbf{0.00\%}\\
& + c &\textbf{1.5909}&\textbf{1.4431}&\textbf{1.1874}&\textbf{0.6696}&\textbf{49.17\%}&0.0223&0.0191&0.0163&0.0115&-1.50\%\\
\hline 
\multirow{3}{4em}{10}
&ORB-SLAM2&0.9651&0.8676&\textbf{0.7517}&0.4227&0.00\%&0.0206&0.0150&\textbf{0.0116}&0.0141&0.00\%\\
& + c &\textbf{0.9190}&\textbf{0.8497}&0.8099&\textbf{0.3501}&\textbf{4.78\%}&\textbf{0.0200}&0.0150&0.0120&\textbf{0.0133}&\textbf{2.76\%}\\
\hline
\end{tabular}
\end{center}
\caption{Experimental results on stereo KITTI Visual Odometry dataset \cite{kitti}.  The best performed methods are highlighted by boldface.}
\label{tab: kitti}
\end{table*}

\subsubsection{Time complexity}
The main time consuming module in our proposed system is weakly supervised semantic segmentation including CRF refinement.  The semantic segmentation module takes RGB images of fixed size of 321 \(\times \) 321 as inputs, and it takes about 1.27 seconds to process 1 image.

\section{Conclusion}
\label{sec:conclusion}
In this paper, we propose a movable object aware vSLAM system modified from ORB-SLAM2 \cite{murORB2}.  We adopt a novel weakly supervised semantic segmentation system \cite{sun2019fully} to leverage the power of deep semantic segmentation CNN, while avoid requiring expensive annotations for training.  Thus, the proposed system is much more applicable and adaptable in new environments compared with other deep learning based approaches.  The color images are passed to the semantic segmentation system to generate pixel-wise categorization masks from which a binary mask is obtained to indicate all the pixels belong to predefined movable object regions.  We modify ORB-SLAM2 so that no feature points fall into the detected movable object regions.  When depth images are available, we incorporate them into the conditional random field (CRF) refinement of the segmentation mask to further improve the performance.  Experimental results on TUM RGB-D and stereo KITTI dataset demonstrate that our approach significantly improves the ORB-SLAM2 in various challenging scenarios, and achieves the state-of-the-art performance in many cases.  To the best of our knowledge, this is the first work that adopts weakly supervised semantic segmentation CNN for dynamic objects aware vSLAM.







\end{document}